# Design, actuation, and functionalization of untethered soft magnetic robots with life-like motions: A review

Jiaqi Miao[1,2]*, Siqi Sun[3]

[1] Department of Mechanical Engineering, The University of Hong Kong, Hong Kong 999077, China
[2] Department of Mechanical Engineering, City University of Hong Kong, Hong Kong 999077, China
[3] Department of Biomedical Engineering, City University of Hong Kong, Hong Kong 999077, China

* Correspondence: jqmiao@connect.hku.hk

**Abstract:** Soft robots have demonstrated superior flexibility and functionality than conventional rigid robots. These versatile devices can respond to a wide range of external stimuli (including light, magnetic field, heat, electric field, etc.), and can perform sophisticated tasks. Notably, soft magnetic robots exhibit unparalleled advantages over numerous soft robots (such as untethered control, rapid response, and high safety), and have made remarkable progress in small-scale manipulation tasks and biomedical applications. Despite the promising potential, soft magnetic robots are still in their infancy and require significant advancements in terms of fabrication, design principles, and functional development to be viable for real-world applications. Recent progress shows that bionics can serve as an effective tool for developing soft robots. In light of this, the review is presented with two main goals: (i) exploring how innovative bioinspired strategies can revolutionize the design and actuation of soft magnetic robots to realize various life-like motions; (ii) examining how these bionic systems could benefit practical applications in small-scale solid/liquid manipulation and therapeutic/diagnostic-related biomedical fields.

## 1. Introduction

Soft robotics is an emerging field centered on the design of flexible machines made from soft materials, and has gained significant attention in recent years [1–6]. Instead of relying on rigid components, soft robots can bend, deform and stretch, allowing them to revolutionize typical robotic tasks, including dexterous manipulation and locomotion. Within soft robotics, one crucial branch is the soft magnetic robot. These robots employ magnetic fields to induce movement [7–10], and can generate considerable forces relative to their size and weight, making them ideal for deployment as small-scale machines [11]. Notably, compared with other actuation principles (e.g., light [12], heat [13], electricity [14], and pneumatic/hydraulic actuation [15–17]), soft magnetic robots present numerous benefits, including wireless actuation, high safety, and fast response. Drawing inspiration from natural organisms, scientists have developed these robots with multiple modes of locomotion [18,19], which enable them to navigate complex environments and perform various tasks, such as small-scale manipulation [20,21], biomedical applications [22–25], microfluidics [26,27], intelligent sensors [28,29], and human-machine interaction [30].

The development of soft magnetic robots is a complex and interdisciplinary process that involves materials selection, actuation, and functionalization. The design of the robot [7,8,31], which encompasses the choice of magnetic materials and geometry, as well as fabrication methods, is critical as it determines the robot's performance, efficiency, and adaptability. The actuation of soft machines is based on the controllable direction and strength of magnetic fields, which can be achieved by using permanent magnetic systems and electromagnetic actuation systems [32–34]. By elaborately controlling the magnetic fields, engineers endow these soft robots with rich life-like motions. The functionalization of magnetic robots is directly related to their actual applications, therefore, has become a hot research topic. Soft robots, with only basic response capability, are no longer able to meet the needs of specific tasks. To overcome this limitation, some functionalization methods have been developed: (i) developing materials with biocompatibility, biodegradability, or multi-responsive abilities [35,36]; (ii) adopting suitable design strategies (such as loading drugs/stem cells/genes… through coating, absorption or embedding) [24]; (iii) switching locomotion modes (based on changed magnetic fields) to adapt to complex environments [18]. All these efforts have greatly impacted the real-world applications of soft magnetic robots.

Some fundamental aspects, from robotic applications of soft actuators [4,11,37–40] to magnetic robots' materials, design, actuation, and applications [7-9,31,33,34,41–43], are well introduced in previous tutorial reviews. Also, some works summarize the development of magnetic robots under some specific themes based on novel perspectives, such as microscale [44,45], shape morphing property [46], biohybrid design [10,47], swarm control [48], and biomedical applications [22,44,45,49,50], which also provide thought-provoking viewpoints. By contrast, this review aims to present a comprehensive overview with a focus on soft magnetic robots' life-like locomotion modes or behaviors [18,51–54] (rolling/tumbling, crawling/walking, jumping, swimming/propulsion, and swarm/assembly) and the developed applications, especially in small-scale solid/liquid manipulation [20,21] and therapeutic/diagnostic-related biomedical applications [55,56] (**Fig. 1**). Specifically, this review paper is organized as follows. First, we discuss the materials employed in the soft magnetic robots and the main magnetic actuation techniques in Section 2. Next, theoretical studies, modeling, and various life-like locomotion modes of soft magnetic robots are further examined in Section 3. Then, it is followed by the introduction of primary applications of soft magnetic robots (Section 4), in particular, small-scale manipulation tasks and biomedical applications. Finally, conclusions and perspectives on soft magnetic robots are given in Section 5.

## 2. Materials, Fabrication, and Actuation Methods

Soft magnetic robots, with adaptive bodies and inherent magnetic responsiveness, have gained great attention in the aspect of miniaturization (ranging from micro/nano-scale to centimeter-scale [18,57–59]). It is feasible to achieve the controllable magnetic field in a small range, while also providing solutions to overcome limitations of conventional miniature robots, e.g., complex design and control, and low adaptability of locomotion. To respond to different external magnetic fields and achieve various functionalities, the design and fabrication of soft robots need to be carefully planned. In Section 2.1, we provide an overview of the fabrication principles of soft magnetic machines, from embedding a single hard magnet to precisely programming the local design. Selecting the appropriate magnetic actuation methods is necessary to attain a broad range of motions and functions of soft robots. As a result, we then summarize several common types of magnetic fields and demonstrate their production by implementing specific devices in Section 2.2.

### 2.1 Magnetic Materials and Fabrication Methods of Soft Magnetic Robots

Soft robots can obtain magnetism from two primary sources: hard magnets and magnetic micro/nano-particles. Hard magnets can be embedded into the soft body to form highly concentrated and localized magnetic moments [60,61], and actuated by an external magnetic field. Alternatively, magnetic micro/nano-particles can be dispersed throughout other soft materials to form distributed magnetic moments [18,21,35,62].

As shown in **Fig. 2a**, soft robots made of hard magnets are usually designed with internal cavities to house the magnet, forming an integrated structure. The embedded magnet enables soft materials to respond to external magnetic fields, allowing for force-dominated translation, torque-dominated rotation, or a combination of these two effects. Moreover, placing magnets with specific orientations inside soft materials can compress and deform them under the actuation of the magnetic field (**Fig. 2b**). These soft robots, with embedded hard magnet(s), are uniformly classified as discrete systems.

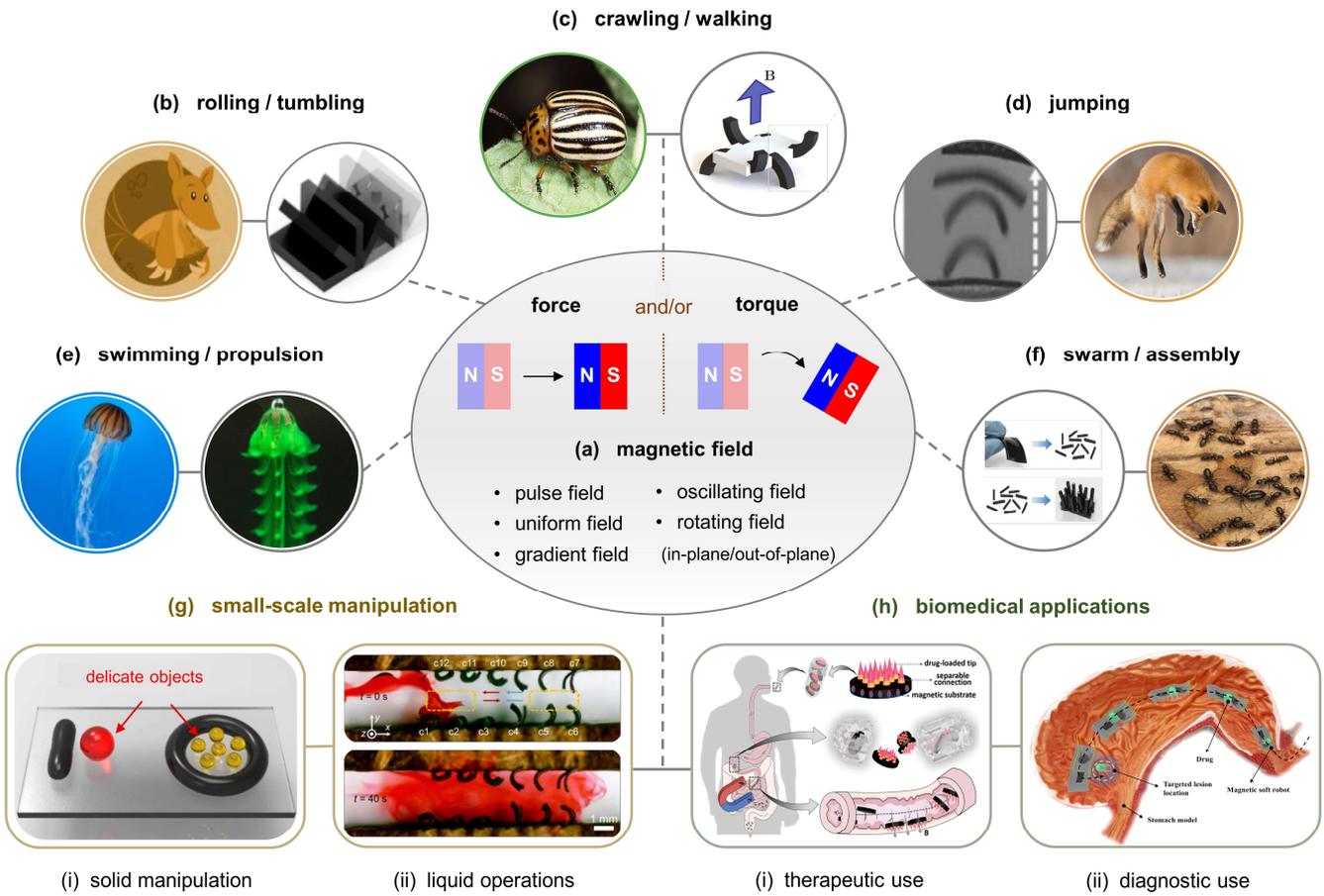

**Figure 1.** Illustration of the review structure. (**a**) Several magnetic fields. (**b**)-(**f**) Life-like motions of soft magnetic robots (with permissions from ref. [18,51–54], Copyright © 2018, Springer Nature; Copyright © 2020, MDPI; Copyright © 2020, Elsevier; Copyright © 2019, Springer Nature; Copyright © 2021, Wiley-VCH). (**g**) Small-scale manipulation of (i) solids (with permission from ref. [20], Copyright © 2020, National Academy of Sciences) and (ii) liquids (with permission from ref. [21], Copyright © 2020, American Association for the Advancement of Science). (**h**) Biomedical applications for (i) therapeutic use (with permission from ref. [55], Copyright © 2021, Wiley-VCH) and (ii) diagnostic use (with permission from ref. [56], Copyright © 2023, Elsevier).

For the dispersed magnetic particles in soft robots (i.e., continuous systems), it is required to consider the magnetization characteristics to choose appropriate fabrication methods. Magnetic materials are usually divided into two categories: soft-magnetic and hard-magnetic. As illustrated in **Fig. 2c(i)**, soft-magnetic materials are characterized by high saturation magnetization ($M_s$), low coercivity ($H_c$), and low remanence ($M_r$), forming narrow hysteresis curves. Conversely, hard-magnetic materials are characterized by large hysteresis curves because of the high $H_c$ and $M_r$ (**Fig. 2c(ii)**). A special case is that the materials have no hysteresis and rapidly saturate under a low magnetic field, which is called superparamagnetic materials. The difference in characteristics also determines their different fabrication and magnetization strategies. For the soft robot comprised of soft-magnetic materials (e.g., iron particles), the external magnetic field is applied to assist the arrangement of magnetic particles in the unsolidified composite matrix), as depicted in **Fig. 2c(iii)**. The robot holds a fixed magnetic arrangement along the magnetic field line after solidification. Even though this arrangement does not truly have a specific magnetization direction, it can respond to the magnetic field in a stable and controllable manner. Thus, several fabrication methods have been developed based on this principle, such as molding assisted with magnetic arrangement [63,64] and template-free magnetic assembly [65–67]. For the soft robot comprised of hard-magnetic materials, e.g., neodymium-iron-boron (NdFeB) particles, micro/nano-particles are first evenly distributed in the soft material, then solidified, and finally magnetized under a strong magnetic field (**Fig. 2c(iv)**). The magnetized soft robots have a strong magnetism and, like a hard magnet, have their specific magnetization directions. Therefore, this fabrication method is also welcomed in soft magnetic robots [68,69].

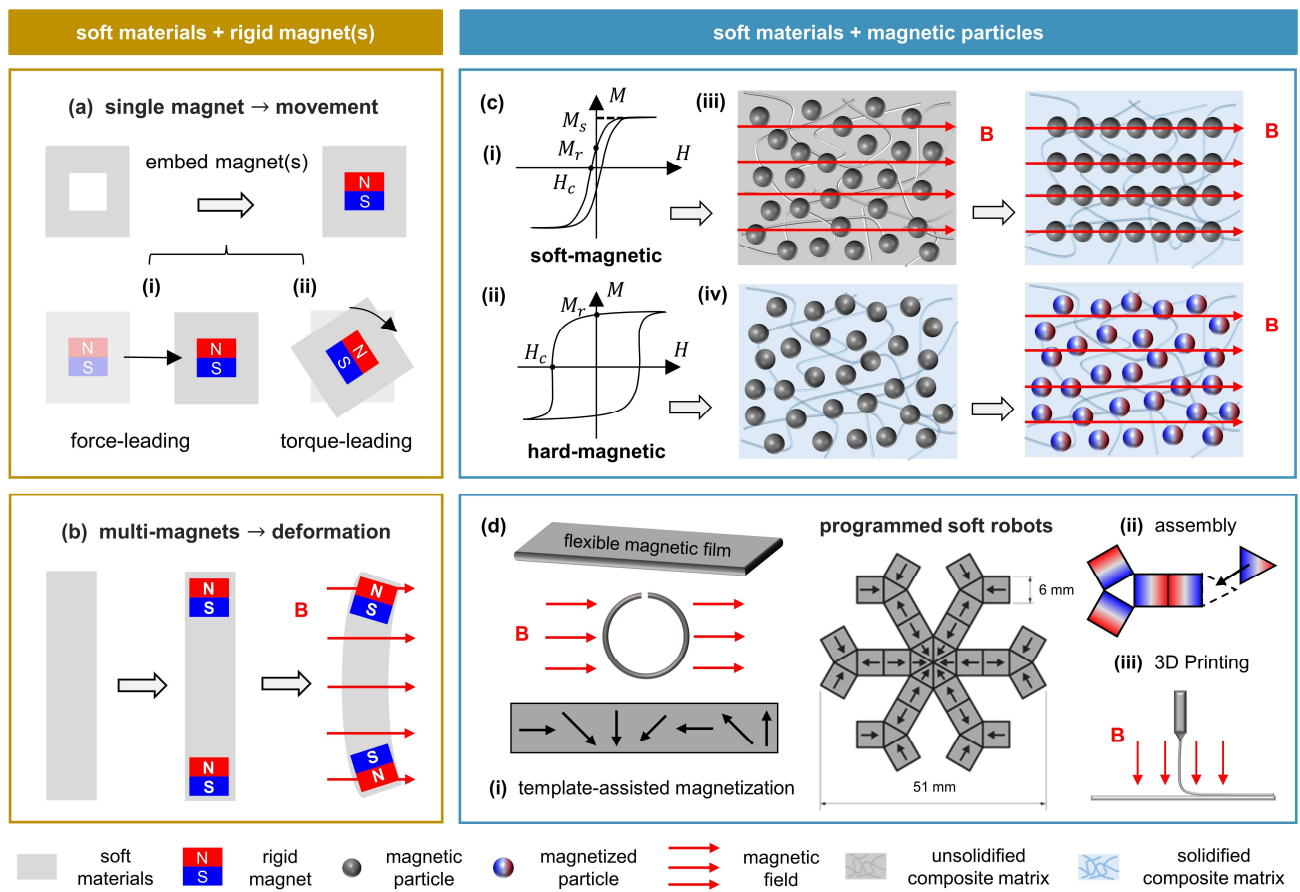

**Figure 2.** (**a**) Illustration of the soft robot embedded with a single magnet and its magnetic-actuated translation and rotation. (**b**) Illustration of the soft robot embedded with multiple magnets and its magnetic-actuated deformation. (**c**) The magnetization characteristics of (i) soft-magnetic materials and (ii) hard-magnetic materials; (iii)-(iv) the corresponding magnetization strategies. (**d**) different magnetization methods of designing programmable soft magnetic robots: (i) template-assisted, (ii) assembly, and (iii) 3D printing.

Magnetically responsive soft robots are mostly composed of magnetic particles and soft materials. A commonly used type of soft materials (with embedded magnetic microparticles or micro-magnets) is silicone, e.g., PDMS and Ecoflex. However, with the development of materials science, the introduction of smart materials brings exciting opportunities for soft magnetic robots [39,70]. Here we briefly enumerate new opportunities brought by the introduction of novel smart materials for soft magnetic robots.

**(i) Shape-memory polymer (SMP):** Considered to be a typical type of smart material, SMPs can return from a deformed state (temporary shape) to their original shape (permanent) when inducing an external stimulus (e.g., temperature change) [71,72]. Currently, SMP-based actuators are endowed with remote controllability and built as multi-stimuli-controlled intelligent micromachines [73]. For instance, SMPs have been introduced into the design of hingeless magnetic origami actuators, combined with either soft magnetic particles [74] or hard magnetic particles [75]. Combining photothermal heating and magnetic actuation, a shape memory polymer matrix with dispersed magnetic microparticles presents rich reconfigurable motions as a magnetic grabber, snapper, or scroll, which can be further integrated with on-board sensing modules.

**(ii) Hydrogel:** Hydrogel, the soft material that has been extensively studied, has been utilized to construct magnetic robots/actuators in the forms of artificial muscles [76], cilia/pillars [77,78], catheters [79], grippers

[36], microrobots [80], and microvalves [81]. The biggest advantages of hydrogels are their tunable mechanical or responsive properties, and especially, their excellent biocompatibility/biodegradability, which enable them to be the preferred choice in biomedical applications. When the magnetic-responsive ability is added, their functionality is anticipated to be further improved and extended [82,83].

**(iii) Liquid crystal elastomer (LCE):** With the combination of rubber elasticity and the orientational order of liquid crystals, LCEs show unique properties such as soft elasticity, thermal actuation, and anisotropic swelling, making themselves suitable for stimuli-responsive actuators [84,85]. By integrating LCEs and magnetic microparticles into monolithic structures, researchers demonstrate the adaptive behaviors of miniature soft robots, including locomotion mode transition in hot liquids, the vine-plant-inspired filament's twining around a hot needle, and a light-switchable magnetic spring [86]. Such a strategy, which exploits the respective advantages of the two materials without compromising their independent stimuli-responsiveness, greatly excites the environment interaction capability of robots.

**(iv) Liquid metal (LM):** With high electrical conductivity, extraordinary fluidity, and biocompatibility, LM has emerged as a promising class of functional materials to serve as actuators, flexible circuits, and biodevices [87,88]. Embedded with LM channels (replacing traditional solid metal coils), the soft elastomer can be designed as a simple, stretchable, durable, and programmable centimeter-scale electromagnetic actuator [89]. The proposed soft actuator, under a low driving voltage, can be maneuverable by the Lorentz force to serve as a dynamic flower actuator, a swinging fish-tail-like actuator, etc. Different from conventional motors, these soft actuators are readily miniaturized with the proposed simple and frugal fabrication method.

Recently, a prominent field is the programmable design of soft magnetic robots, which have well-designed local magnetization directions and can exhibit abundant deformations and motions under magnetic actuation [90–92]. **Fig. 2d** depicts the three magnetic shape-programming techniques that are commonly adopted. As illustrated in **Fig. 2d(i)**, the first method involves molding the soft robots and magnetizing them with the aid of a template. The composed materials are hard-magnetic, and the fabricated soft magnetic robots can be easily programmed as expected, leading to excellent movability and controllability [18,62,92]. The second programming strategy entails assembling the magnetized units manually, according to the desired design (**Fig. 2d(ii)**). The magnetic building blocks can combine into diverse magnetization patterns, and then form various 3D shapes under the actuation of magnetic fields, which can be highly sophisticated, but also laborious [93–95]. The third magnetic programming design approach involves the combination of 3D printing technologies (**Fig. 2d(iii)**). For example, the extrusion-based 3D printing technology can reorient hard-magnetic particles with the assistance of weak magnetic fields, and further control the magnetization patterns [57]. Additionally, light-based 3D printing technology has been used to fabricate both soft-magnetic and hard-magnetic materials. During the selective curing process, magnetic fields are applied to align the magnetized/magnetizable particles and program the robots [90,96,97].

The programmable design of soft magnetic robots contributes to the realization of rich biomimetic motions (emphatically introduced in Section 3), and wonderful functional origami [98–102]. An impressive example is that by combining the 3D printing technology, soft ferromagnetic robots have been designed with complex shape changes, which may benefit reconfigurable soft electronics, jumping mechanical metamaterial, and multi-modal robots [57]. Based on the extension of the Kresling pattern, Wu et al. propose an origami-based robotic arm that enables sophisticated motions, e.g., stretching, contracting, reconfigurable bending, and multi-axis twisting [103]. In subsequent studies, soft origami millirobots with similar designs are developed as an untethered type, and show strong locomotion abilities (crawling, rolling, flipping, and steering capabilities) and potential for biomedical applications (the folding/unfolding-based pumping mechanism for controlled drug delivery) [104,105]. A similar idea, using magnetic origami robots for *in vivo* biomedical applications, originated in the work of Miyashita et al [106]; however, the magnetically controlled locomotion at that time was not complicated, and origami design was not utilized to develop rich motions. Its highlight is to show how a lightweight, foldable, movable, and degradable magnetic millirobot may play crucial roles in future biomedical applications. To sum up, the design of soft magnetic robots requires considering target functions/applications and matching the right materials and fabrication methods. We refer the reader to reviews that focus on materials, fabrication, and design of soft robots [31,107,108].

## 2.2 Magnetic Actuation Methods and Devices

Multiple standardized magnetic fields have been proposed for soft robots, including pulsed field, uniform field, gradient field, oscillating field, and rotating field, which pave the way to plenty of locomotion modes and actual applications. **Fig. 3a** summarizes the characteristics of each standardized magnetic field. The pulsed field is essentially a regular switching-on/off field (**Fig. 3a(i)**); the uniform field has a uniformly distributed field strength with the same magnetic field direction (**Fig. 3a(ii)**); the gradient field has an increased/decreased field strength (**Fig. 3a(iii)**); the oscillating field refers to the field that oscillates back and forth to change the magnetic field directions (**Fig. 3a(iv)**); the rotating field indicates the field that has a periodic in-plane or out-of-plane rotation to change the magnetic field directions (**Fig. 3a(v)**). These standardized magnetic fields are achieved mainly through two categories of devices: permanent magnet systems and electromagnet systems, which will be discussed below, respectively.

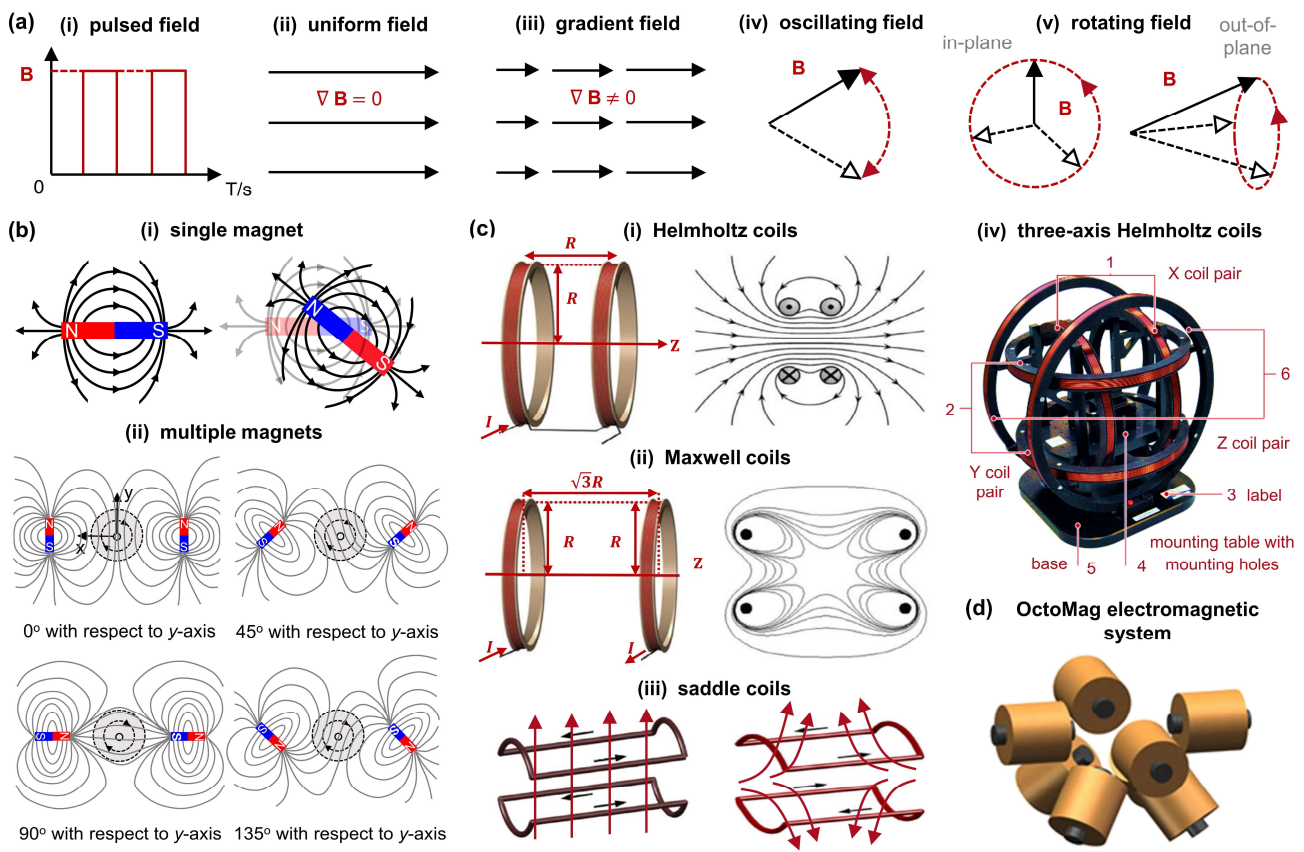

**Figure 3.** (**a**) Several standardized types of magnetic fields. (**b**) Permanent magnet systems: (i) a single magnet; (ii) multiple magnets. (**c**) Orthogonal paired coils: (i) Helmholtz coils; (ii) Maxwell coils (with permission from ref. [122], Copyright © 2018, Springer Nature); (iii) saddle coils (with permission from ref. [123], Copyright © 2010, Elsevier); (iv) three-axis Helmholtz coils. (**d**) Distributed electromagnets: OctoMag electromagnetic actuation system (with permission from ref. [8], Copyright © 2022, Elsevier).

Permanent magnets are readily available, highly compact, and with negligible heat loss, which can produce a stronger magnetic field than electromagnets. Thus, they are ideal for actuating soft magnetic robots.

As shown in **Fig. 3b(i)**, a single magnet can be regarded as the simplest permanent magnet system. With the changed postures, it could control the translation and rotation of soft robots by applying dynamic forces or torques [109–111]. Although the manual operation is feasible, it requires and highly depends on the operator's experience. To enhance operating accuracy and avoid environmental hazards for operators, permanent magnets are connected to the multi-degree-of-freedom robotic arms to provide remote and automatic control [112–114]. However, the tricky problem is that a single permanent magnet cannot create a uniform magnetic field. As a possible solution, researchers suggest employing multiple magnets to impair the field gradient and generate a uniform magnetic field for torque actuation. For instance, a magnetic system that has two synchronized rotating permanent magnets can produce a uniformly rotating magnetic field [115], as displayed in **Fig. 3b(ii)**. According to specific purposes, multi-magnet systems can be devised into numerous arrangements to fulfill the requirements of diverse magnetic field control [116–129].

Electromagnet systems consist of an inserted magnetic core and external solenoid, which provide controllable strength and direction of magnetic fields [120,121]. These devices are popular because they do not have the complex coupled magnetic field like permanent magnets, and are adjustable to the size and shape of the workspace on demand. Here, we categorize the electromagnet systems into orthogonal paired coils and distributed electromagnets for separate discussions.

Helmholtz coils (**Fig. 3c(i)**), consisting of two coils arranged at a distance equal to their radius with an electric current in the same direction, are widely used to produce uniform magnetic fields [122]. Maxwell coils (**Fig. 3c(ii)**), another classic layout, have a similar configuration to Helmholtz coils, but differ in the distance of two coils (changed to $\sqrt{3}R$, $R$ is the radius of the coils) and the electric current direction (Maxwell coils are energized by reverse currents), thereby used to generate gradient magnetic fields [122]. As in **Fig. 3c(iii)**, to reduce space occupation and heat loss, the paired coils have been developed as curved-shape saddle coils, which generate a uniform or gradient magnetic field by applying the same or opposite currents [123]. In practical applications, paired coils are mainly used to obtain uniform magnetic fields. Currently, the three-axis Helmholtz coil system (**Fig. 3c(iv)**) is most widely adopted to attain uniform magnetic fields towards any direction in three-dimensional space [124–127]. This coil system can generate a uniform magnetic field near the center. However, the available working space is limited by the innermost coils, making it suitable for small magnetic robots. In addition to basic triaxial arrangements, Helmholtz coils can be configured as two pairs to provide an in-plane uniform magnetic field [128]. They may also be fabricated in a square shape to enlarge the operating space [129]. Moreover, Helmholtz coils can be incorporated with Maxwell or saddle coils to realize additional control goals [123,130,131].

Distributed electromagnets, with columnar coils pointing towards the workspace, are another common type of electromagnet system, which can relieve tight arrangements and enhance energy efficiency. The expected magnetic field is achieved by superimposing the magnetic fields generated by each electromagnet. The OctoMag system, a well-known example of electromagnet system (**Fig. 3d**), consists of eight electromagnets that can govern the five degrees of freedom (DOFs) of magnetic robots [132]. Its eight electromagnets are grouped into two: the upper group comprises electromagnets inclined at an angle of 45º to their common axis; the lower group has four electromagnets in one plane with an angle of 90º between adjacent magnets. A broad range of designs of such distributed electromagnets is developed to cater to different needs, mainly reflected in the arrangement and number of electromagnets [133–136]. We refer the reader to reviews that focus on magnetic actuation systems for details [34,41].

Besides conventional electromagnet systems' far-field actuation, recent works show that the near-field magnetic actuation by bringing the field source and magnetic actuators in close proximity [20]. Such a driven mode provides higher energy efficiency and compactness, also offers appealing potentials in portable magnetic robotic systems with light weight and low energy consumption [137]. Fan et al. propose reconfigurable ferrofluid droplet robots based on the near-field magnetic actuation [20]. The droplet robots, with the spatiotemporal magnetic field control, exhibit on-demand splitting and merging for delivering cargo, and can be independently controlled. Further, when microscale thickness flexible planar coils are adopted in the near-field actuation, soft actuators are capable of lifting, tilting, pulling, and grasping that are independently addressable. It is worth noting that the ultrathin and lightweight actuators work with a high frequency but low energy consumption; meanwhile, the miniaturization of the actuation system makes the whole system extremely portable.

## 3. Life-like Locomotion of Soft Magnetic Robots

A comprehensive analysis of the design, fabrication, and actuation methods of soft magnetic robots enables us to develop their biomimetic motions and acquire generalized strategies. Here, we focus on the multimodal locomotion of miniature soft magnetic robots, emphasizing how artificial systems can replicate life-like motions/behaviors inspired by nature and benefit a wide spectrum of engineering fields. To this end, we first briefly introduce the theoretical and modeling parts of magnetically controlled robots, which are the basis for designing, optimizing, and evaluating soft robots' various motions. Ensuingly, five common categories of bioinspired motions (containing rolling/tumbling, crawling/walking, jumping, swimming/propulsion, and swarm/assembly) are respectively analyzed, which involve various environments (air, land, and water) and objects ranging from individuals to swarms.

### 3.1 Theory and Modeling

Magnetic actuated is conducted by applying forces and/or torques on magnetic robots. Because there is no current in the operating area, the quasi-static magnetic field used for magnetic manipulation can be described by Maxwell's equations:

$$\nabla \cdot \mathbf{B} = 0 \tag{1}$$

$$\nabla \times \mathbf{B} = 0 \tag{2}$$

where **B** is the magnetic-vector field and $\nabla$ is the gradient operator. When placed in a magnetic field **B**, the magnetic torque **τ** acting on the magnetic robot with dipole moment **m** is described as:

$$\boldsymbol{\tau} = \mathbf{m} \times \mathbf{B} = \begin{bmatrix} 0 & B_z & -B_y \\ -B_z & 0 & B_x \\ B_y & -B_x & 0 \end{bmatrix} \begin{bmatrix} m_x \\ m_y \\ m_z \end{bmatrix} \tag{3}$$

Meanwhile, when the magnetic robot is in a non-uniform magnetic field, the produced magnetic force **F** can be expressed as:

$$\mathbf{F} = (\mathbf{m} \cdot \nabla)\mathbf{B} = \begin{bmatrix} \dfrac{\partial B_x}{\partial x} & \dfrac{\partial B_x}{\partial y} & \dfrac{\partial B_x}{\partial z} \\ \dfrac{\partial B_x}{\partial y} & \dfrac{\partial B_y}{\partial y} & \dfrac{\partial B_y}{\partial z} \\ \dfrac{\partial B_x}{\partial z} & \dfrac{\partial B_y}{\partial z} & -\left(\dfrac{\partial B_x}{\partial x} + \dfrac{\partial B_y}{\partial y}\right) \end{bmatrix} \begin{bmatrix} m_x \\ m_y \\ m_z \end{bmatrix} \tag{4}$$

Equations (3) and (4) abstract the magnetic robot into a dipole. The magnetic force and torque can act simultaneously or separately, that is, up to six degrees of freedom (6 DOFs) can be executed on the robot. However, a uniformly magnetized robot with axial symmetry has at most 5 DOFs, because the symmetry around the magnetization direction reduces one rotational DOF. As for the magnetic field, it contains 8 DOFs, i.e., 3-DOF magnetic field components and 5-DOF field gradient components.

As mentioned before, the magnetic field can be produced by permanent magnets or electromagnets. For the permanent magnet, the produced magnetic field depends on the magnet's size and shape, which can be conveniently calculated by the dipole model. At the region of interest, the generated magnetic field $\mathbf{B_r}$ is:

$$\mathbf{B_r} = \frac{\mu_0}{4\pi \|\mathbf{r}\|^3} \left( \frac{3\mathbf{r}\mathbf{r}^T}{\|\mathbf{r}\|^2} - \mathbf{I} \right) \mathbf{M} \tag{5}$$

where $\mu_0$ is the magnetic permeability of free space, $\mathbf{r}$ is the robot's position vector relative to the driving magnet, $\mathbf{I}$ is the identity matrix, and $\mathbf{M}$ is the dipole moment of the driving magnet. By submitting Equation (5) to Equation (4), the exerted force $\mathbf{F_d}$ on the magnetic robot is given by:

$$\mathbf{F_r} = \frac{3\mu_0}{4\pi \|\mathbf{r}\|^4} \left[ \frac{\mathbf{M}\mathbf{r}^T}{\|\mathbf{r}\|} + \frac{\mathbf{r}\mathbf{M}^T}{\|\mathbf{r}\|} - \left( \frac{5\mathbf{r}\mathbf{r}^T}{\|\mathbf{r}\|^2} - \mathbf{I} \right) \frac{\mathbf{M}^T\mathbf{r}}{\|\mathbf{r}\|} \right] \mathbf{m} \tag{6}$$

Yet, such an approximation will fail when the magnetic robot is actuated at a short range. At that time, other methods need to be considered. The electromagnet shows a more complex field distribution, especially considering the inserted iron cores that are used to increase the magnetic field intensity. Therefore, the dipole approximation is no longer appropriate. To solve that, some other effective evaluation methods have been proposed, e.g., the Biot–Savart law-based mathematical model [138], the fitted dipole model [132], etc. More information about the theoretical analysis on magnetic methods in robotics can be found in some specialized reviews/chapters [34,139].

Moreover, it is noted that there are also very strong modelling activities for soft magnetic robots, which provide useful guidance for the design of robots to a certain extent. With the input of relevant parameters, simulation results will provide us with room for optimization. For instance, by developing codes that are compatible with Abaqus, Kim et al. have shown the great role of modeling in diverse soft robot designs [57]. When driven at a fast speed, polymer-based soft actuators could experience mechanical instabilities, the modeling is highly welcomed in the design of such magnetic actuators. Combining simulations and experiments, Wang et al. show that the proposed ultrafast soft-bodied robots can walk, swim, levitate, transport cargo, and trap a living fly [140].

*3.2 Rolling or Tumbling Locomotion*

Rolling or tumbling is an effective locomotion mode used by natural organisms to achieve high moving speeds and escape from danger [141,142]. Due to such advantages, many soft magnetic robots have been developed with rolling or tumbling motion abilities [18,51,68,109,143,144]. Rolling or tumbling motion is commonly activated by a rotating magnetic field and demonstrates high efficiency, which can be applied to soft magnetic robots with different forms and characteristics.

Bi et al. present microscale magnetic tumbling robots (μTUMs) fabricated through a photolithography process [143], as shown in **Fig. 4a(i)**. Magnetic particles in μTUMs can align along different axes under magnetic fields, and the magnetic parts on the two sides of μTUMs can feature diverse shapes. **Fig. 4a(ii)** illustrates that these μTUMs can perform a variety of tumbling motions when actuated by a rotating magnetic field.

Soft magnetic robots can achieve rolling motion when the body is pre-curved into a ring or barrel shape. Hu et al. propose a soft magnetic robot that can achieve multimodal motions [18], one of which is the rolling motion. The programmable design of the robot allows it to maintain a wheel-like shape when subjected to a uniform magnetic field, as depicted in **Fig. 4b(i)**. The rotating magnetic field can drive the soft robot to rapidly roll over a solid substrate. **Fig. 4b(ii)** shows a one-cycle rotation of the robot under a rotating magnetic field with a magnetic field intensity of 18.5 mT. Yeow et al. design a magnetic-responsive robot that is programmed with local magnetization [109]. When an external permanent magnet is brought near, the programmed robot, actuated by the magnetic field, bends into a ring shape (**Fig. 4c(i)**). Then, following the rotating external magnet, the robot (with dimensions of 20 mm × 4 mm) can perform synchronous rotating motion (as seen in **Fig. 4c(ii)**), and reach a speed of 9.9 mm/s (**Fig. 4c(iii)**). Gu et al. introduce a soft magnetic carpet [68], which is fabricated via a molding process and placed under a strong magnetic field for programmable magnetization (**Fig. 4d(i)**). Driven by a relatively strong rotating magnetic field (>60 mT), the applied magnetic torque can overcome the gravity and elastic energy of soft carpets, resulting in the rolling of the soft carpet. As illustrated in **Fig. 4d(ii)**, the whole process undergoes a sequence of curling, rolling, and opening operations.

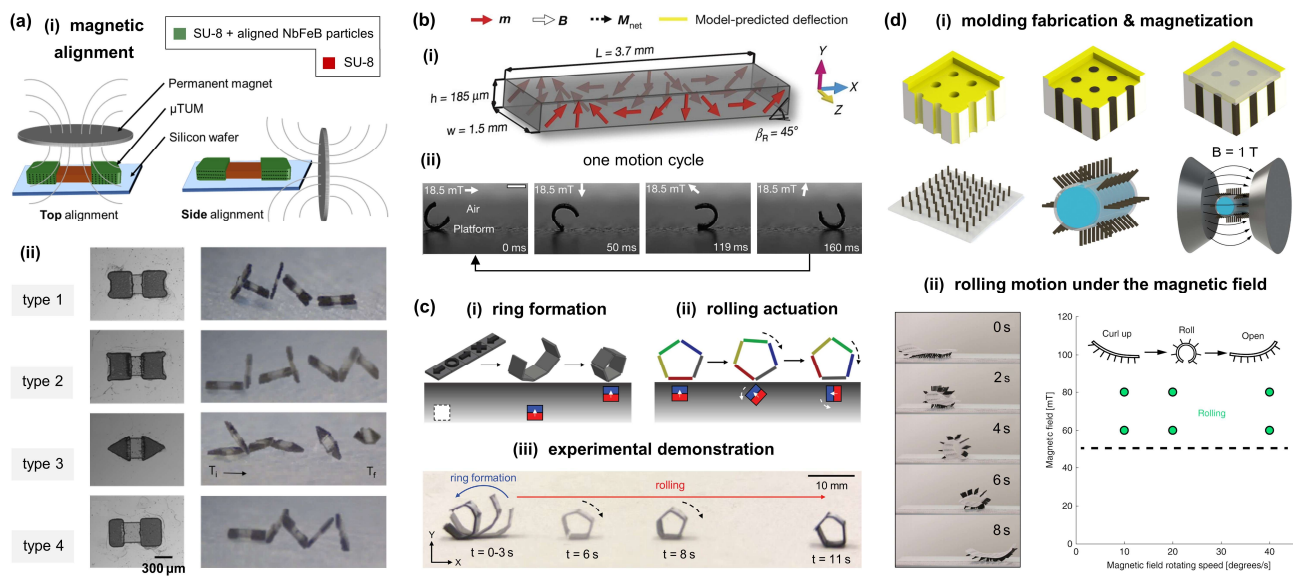

**Figure 4.** (**a**) (i) μTUMs with different magnetic alignment axes. (ii) Tumbling of four types of μTUMs (with permission from ref. [143], Copyright © 2018, MDPI). (**b**) (i) Soft magnetic robot with programmable design. (ii) One-cycle rolling motion (with permission from ref. [18], Copyright © 2018, Springer Nature). (**c**) (i) Ring formation of the magnetic responsive robot under the magnetic actuation. (ii) Illustration of magnetic actuation for rolling motion. (iii) Experiments of the rolling motion (with permission from ref. [109], Copyright © 2022, Wiley-VCH). (**d**) (i) Fabrication and magnetization of the soft magnetic carpet. (ii) Experimental illustrations of the rolling motion (with permission from ref. [68], Copyright © 2020, Springer Nature).

In summary, rotating or tumbling locomotion exhibits a substantial motion ability and can be incorporated into different forms of soft magnetic robots. Furthermore, this technique has widespread versatility in robots at a variety of scales, from micro/nano-robots [145,146] to (sub)millimeter-scale soft machines [18,147].

### 3.3 Crawling or Walking Locomotion

Crawling and walking are two typical terrestrial motions found in numerous natural organisms [148]. The crawling motion is usually accompanied by the organism's body deformation, and with the help of frictional forces at the anchor point, it can obtain the counteractive propulsion force. Legged organisms use walking locomotion to move forward and cross obstacles through rhythmic interactions of multi-legs. These two modes of motion have been widely studied in robotics because of their diversity and superiority in terrestrial movement [4,149–152]. Developing miniature soft magnetic robots that can replicate these unique motion modes requires careful consideration of design/actuation strategies inspired by biology [18,52,64,153–157].

To imitate the snake-like movement, Kim et al. propose a programming magnetic microrobot consisting of four parts with different magnetic arrangement axes [153]. When a homogeneous magnetic field is applied, each part of the robot rotates in different directions and angles to facilitate crawling (**Fig. 5a**). Zhang et al.

present the programmed microscopic magnetic artificial cilia (μMAC) capable of crawling motion [64]. When placed in a uniform magnetic field, the arrayed artificial cilia display different bending states depending on their different magnetic arrangement directions. Consequently, the rotating magnetic field can actuate the μMAC to crawl in a metachronal manner (with a phase difference between neighboring beating cilia), as portrayed in **Fig. 5b**. Robots with a similar multi-legged design can be developed with other locomotion modes. By changing the path of the external magnet, Lu et al. show the discontinuous flap wave mode and continuous inverted pulse mode of the soft millirobot, and demonstrate its movement ability in both dry and wet environments [110]. **Fig. 5c** shows the crawling and walking motions of the soft-bodied magnetic robot [18]. When the robot is confined in a hollow pipe, under a rotating magnetic field, it can produce a longitudinal traveling wave to crawl like a caterpillar (**Fig. 5c**: left side). Inspired by the gait of an inchworm, the soft robot can also present the walking motion by alternately anchoring its front/back end to pull itself forward (**Fig. 5c**: right side). The crawling motions presented above are mainly replicated through programmable designs of soft robots. A different approach is to depend on external magnetic actuation. Joyee et al. develop a soft robot that is designed with anterior and posterior magnetic legs [155]. Its crawling motion is achieved by alternately moving the magnets beneath the legs.

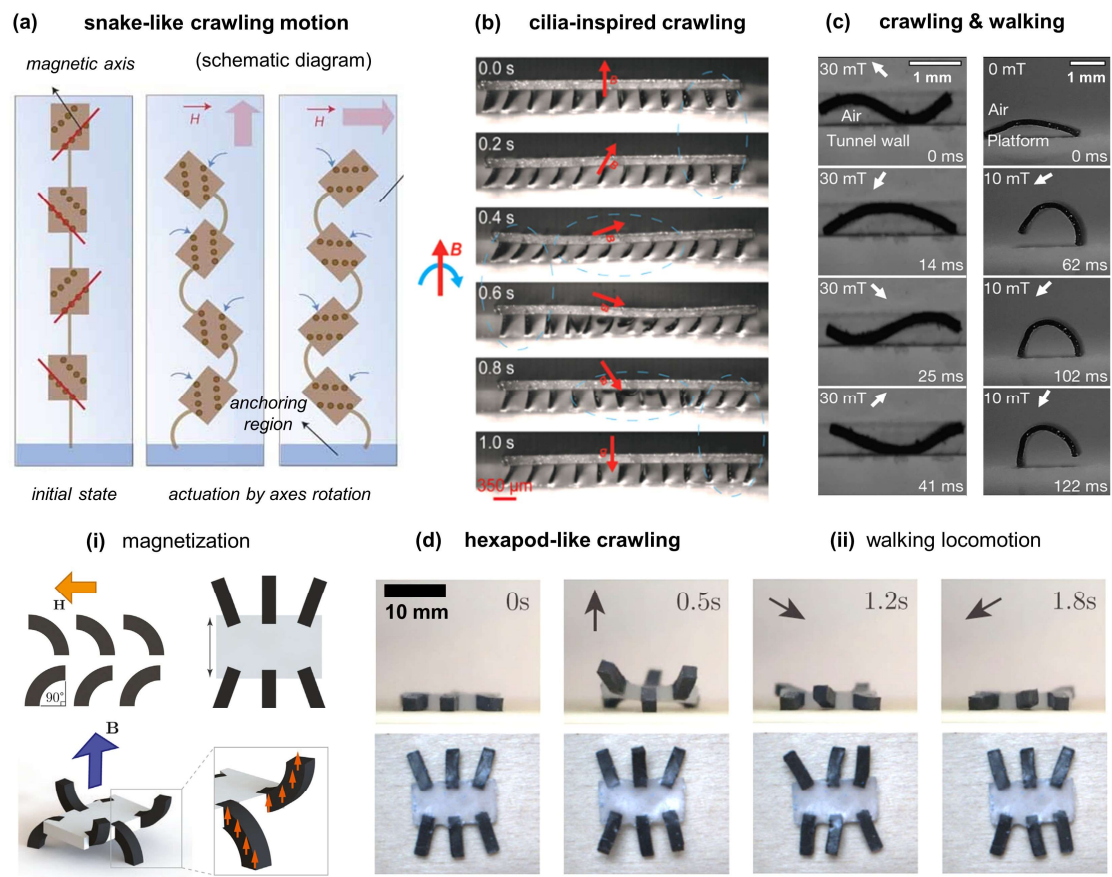

**Figure 5.** (**a**) The snake-like crawling motion of the four-part programmed robot (with permission from ref. [153], Copyright © 2011, Springer Nature). (**b**) The cilia-inspired metachronal crawling motion of the μMAC (with permission from ref. [64], Copyright © 2021, American Chemical Society). (**c**) Crawling and walking motions of the soft-bodied magnetic robot (with permission from ref. [18], Copyright © 2018, Springer Nature). (**d**) Hexapod-like crawling robot: (i) magnetization arrangement; (ii) walking motion (with permission from ref. [52], Copyright © 2020, Elsevier).

A more common walking pattern in nature involves the coordination of multi-legs. Venkiteswaran et al. introduce a multi-legged magnetic robot [52]. Its six legs are divided into two equal groups, and each with different magnetization directions (**Fig. 5d(i)**). The robot has three pair of legs, and each pair is anti-symmetric, which ensure a tripod support at any time of the motion cycle. The legs of the soft magnetic robot can step forward with the provided magnetic torques from a rotating magnetic field. **Fig. 5d(ii)** shows the snapshots of walking locomotion (top row: side view; bottom row: top view).

Crawling and walking exhibit robustness in land locomotion, and their diverse gaits have been thoroughly studied in traditional macro-scale rigid robots [158,159]. Yet, replicating them at a small scale remains a significant challenge. Current strategies for magnetic robots mainly consist of planning external magnetic field control and providing programmable design. The crawling/walking soft robots are relatively easy to implement in the scale range from (sub)millimeters to centimeters. Although there are cases at the micro/nano-scale, complex and meticulous design/fabrication is required [160].

*3.4 Jumping Locomotion*

Aerial locomotion, an important category of movement, is observed in a wide range of organisms and serves as a significant source of inspiration for robot design [161–164]. Yet, soft magnetic robots are known for their portability which implies that they lack an integrated power supply module to support a long-time flight. Thus, their brief aerial flight is usually presented in the form of jumping locomotion. To overcome obstacles and gaps, soft robots have used jumping locomotion by rapidly releasing stored elastic energy [18,165–169]. **Fig. 6** highlights several jumping strategies of soft magnetic robots for further illustration.

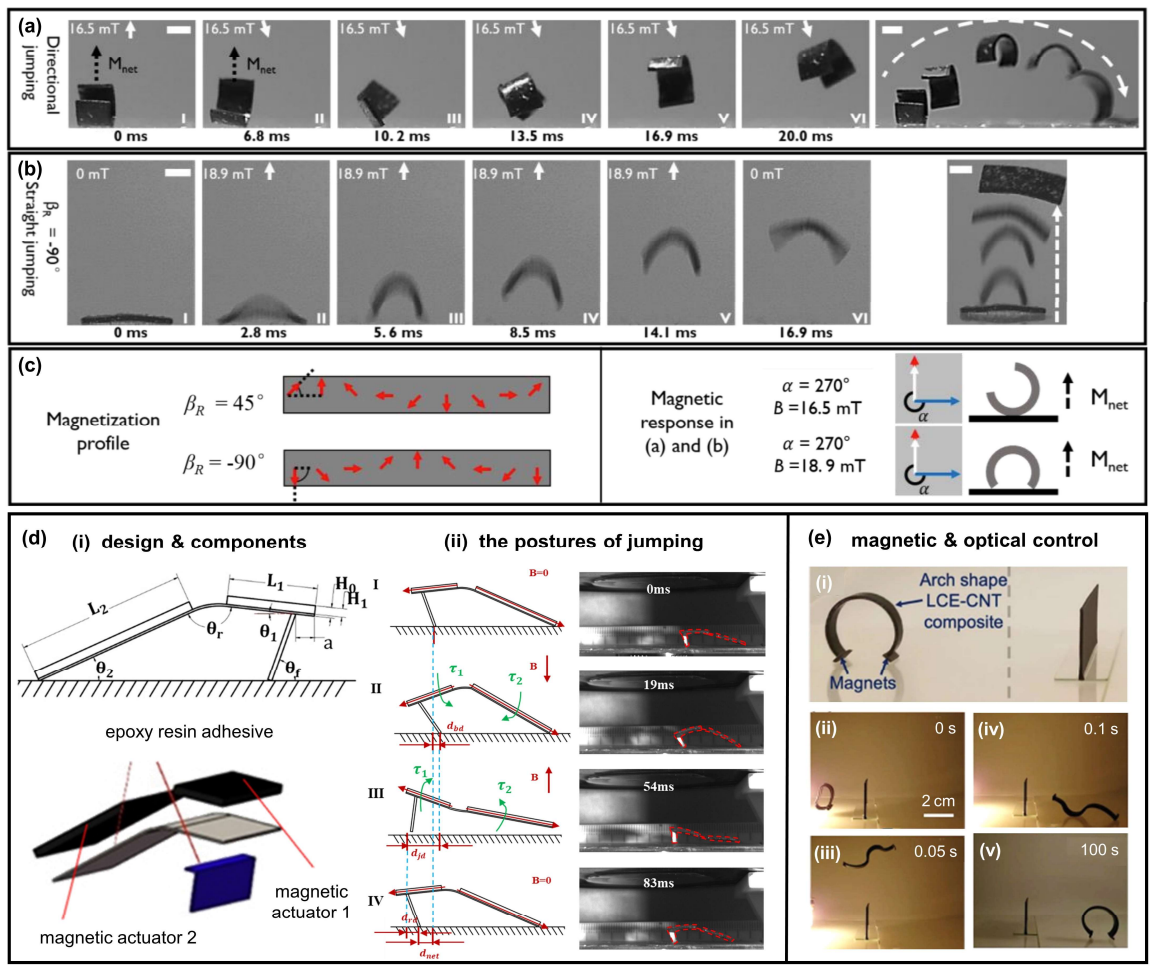

**Figure 6.** (**a**) Illustration of directional jumping of the soft magnetic robot. (**b**) Illustration of straight jumping of the soft magnetic robot. (**c**) Magnetization profiles of two soft robots with different jumping modes (with permission from ref. [18], Copyright © 2018, Springer Nature). (**d**) (i) Design of the soft jumping robot. (ii) Postures of jumping motion (with permission from ref. [167], Copyright © 2023, Springer Nature). (**e**) (i)-(v) Jumping experiments of the robot under magnetic and optical control (with permission from ref. [168], Copyright © 2019, Wiley-VCH).

Hu et al. develop two types of jumping modes of the soft-bodied magnetic robot [18], directional jumping (**Fig. 6a**) and straight jumping (**Fig. 6b**), by designing different magnetization profiles (**Fig. 6c**). For directional jumping, the robot first curls upon the application of a magnetic field, which accumulates elastic energy. Then, the direction of the magnetic field is changed (within a few milliseconds) to actuate the robot's jumping motion (with rotation) towards the expected direction (**Fig. 6a**). This actuation method can adjust the jumping direction according to the principle that the net magnetic moment of the soft robot prefers to align with the imposed magnetic field. For straight jumping (**Fig. 6b**), the soft robot reacts to a sudden upward magnetic field and rapidly jumps up. Subsequently, the magnetic field is immediately removed. Following the jump, the soft robot descends to the ground due to gravitational force.

Soft centimeter-scale magnetic robots can achieve jumping locomotion by designing specific mechanical structures or endowing multi-responsive capability. Inspired by the structure of quadrupeds, Zhong et al. present a magnetic single-leg soft jumping robot (SLSJR) that is composed of a main body part, an elastic leg, and two magnetic actuators (**Fig. 6d(i)**) [167]. In one jumping cycle, there are four stages: (I) initial state - without the external magnetic field; (II) inward bending - when a vertical downward magnetic field is applied, two magnetic actuators, with different magnetization directions, will fold inward in response to the magnetic field; (III) outward extending - a subsequent vertical upward magnetic field actuates the robot to extend outward and raise its front leg to jump; (IV) back to the initial state - after the magnetic field is removed, the robot restores its original state and obtains a horizontal net displacement. The corresponding snapshots of the experiment are displayed in **Fig. 6d(ii)**.

Ahn et al. propose a soft robot capable of jumping like the fly larva, combinedly utilizing magnetic and optical control [168]. The experimental layout is shown in **Fig. 6e(i)**, and the jumping process is displayed in **Fig. 6e(ii)-(v)**. The soft robot first bends into a closed loop by exposing its inner surface to light. With the attracting force between magnets, the robot can maintain a ring shape, even after the light source has been turned off. Then, the outer surface is exposed to light to deform the robot and store elastic energy. As soon as the stored energy reaches the robot's elastic limit, the jumping motion occurs. The soft robot (40 mm × 7 mm × 2 mm) can jump up to 80 mm, while traveling a distance of approximately 130 mm. Although multi-responsive soft robots are not the central theme of this review, it is noteworthy that they offer an alternative means of realizing specific motions and functions. For certain motions, such as jumping, which require the accumulation of energy for instantaneous release, light and chemical actuation are superb power supply options [166,168]. In addition, the levitation of magnetic robots is also worth noting [140,170], as they can stay in the air for a considerable amount of time instead of as short as the jumping motion.

*3.5 Swimming Locomotion or Propulsion*

Aquatic locomotion takes place in the liquid environments surrounding or within organisms. The most effective strategy for designing and operating flexible and efficient artificial swimmers is to derive inspiration from the swimming styles of aquatic animals [171–174]. For soft magnetic robots, they are highly appreciated due to their advantages in biomedical applications (e.g., targeted drug delivery and minimally invasive surgery) [22]. To perform medical operations within the human body, magnetic robots should be small enough. Hence, the changes in fluidic characteristics should be taken into careful consideration since they directly impact the choice of appropriate swimming strategies. Here, we respectively discuss micro/nano-scale and (sub)millimeter-scale magnetic swimmers that are bound by the physical laws governing Stokesian hydrodynamics.

3.5.1 Micro/nano-swimmers

Micro/nano-scale swimming occurs at low Reynolds numbers, where viscous forces dominate motions while inertial forces can be ignored [175,176]. In contrast to macroscopic scenarios, inertial-force-dominated motions can no longer obtain net displacement. To solve this dilemma, researchers have looked to natural microorganisms for inspiration and developed artificial bionic robots that imitate their motility [177]. Cilia and flagella are the main motion units of microorganisms, displaying efficient propulsion abilities at the microscale. Their unique motions offer simple but effective design strategies for low-Reynolds-number swimmers [178].

Artificial cilia are man-made systems imitating the beating modes of natural cilia, ranging from individual cilium to arrays. Hanasoge et al. propose an artificial magnetic cilia system, which reproduces the asymmetric ciliary beating at the individual level [179]. **Fig. 7a** illustrates that when a counterclockwise rotating magnetic field is applied, the cilium first goes through a power stroke (blue line) due to the magnetic actuation. When coming to its elastic limit, the cilium quickly returns to its original position, i.e., recovery stroke (red line), with a different path. This spatial asymmetry creates a non-reciprocal motion mode, allowing the cilium to propel effectively at low Reynolds numbers. The asymmetric characteristics of natural cilia are diverse [180], e.g., orientational asymmetry in three-dimensional (3D) ciliary beating and metachronal coordination in cilia arrays. We refer the reader to artificial cilia reviews for more details [26,27,181,182].

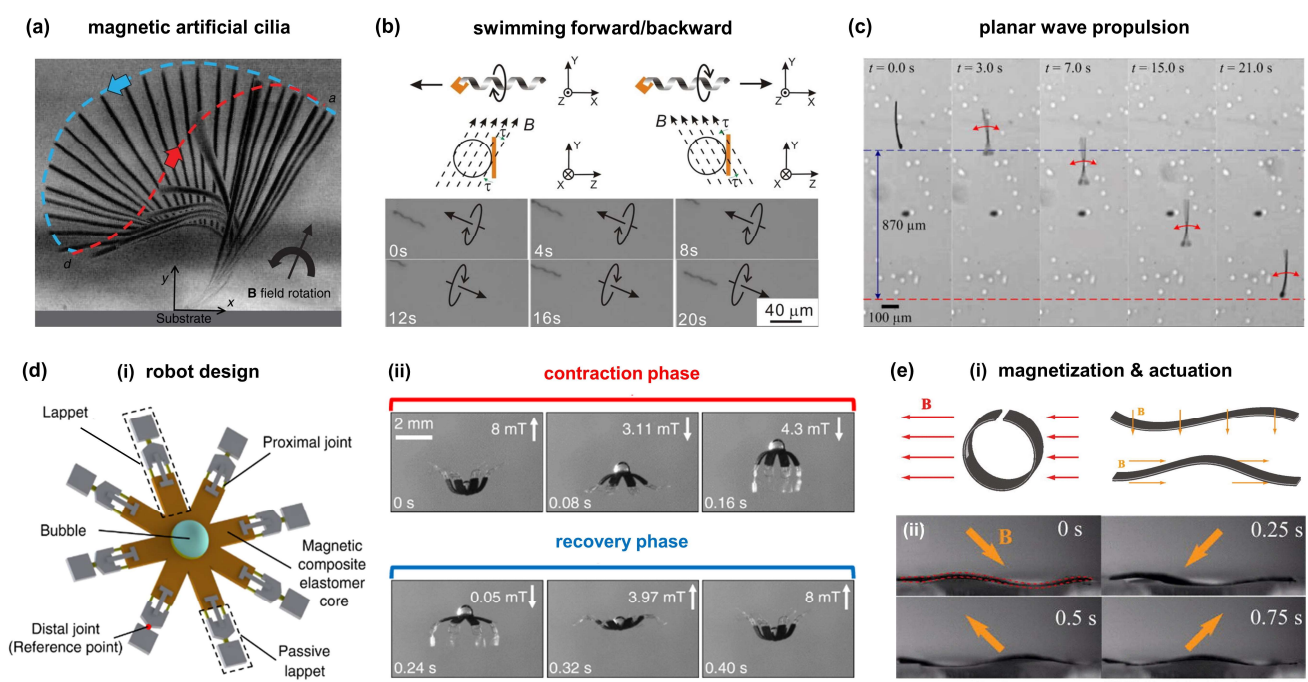

**Figure 7.** (**a**) Asymmetric beating a single magnetic cilium (with permission from ref. [179], Copyright © 2018, Springer Nature). (**b**) The forward/backward swimming of artificial bacterial flagella (ABF) swimmer (with permission from ref. [183], Copyright © 2009, AIP Publishing). (**c**) Illustration of the planar wave propulsion of the sperm-inspired magnetic swimmer (with permission from ref. [190], Copyright © 2014, AIP Publishing). (**d**) (i) Design of the jellyfish-like soft robot; (ii) Contraction and recovery phase of the soft robot (with permission from ref. [53], Copyright © 2019, Springer Nature). (**e**) (i) Magnetization and actuation of the soft magnetic robot; (ii) Swimming under the control of the magnetic field (with permission from ref. [92], Copyright © 2014, AIP Publishing).

Zhang et al. fabricate the magnetically responsive artificial bacterial flagella (ABF) inspired by bacterial motion [183]. The ABF swimmer rotates its body in response to rotating magnetic fields to propel itself, utilizing the same mechanism as natural bacteria [184]. Because of the chiral asymmetric structure, the rotating motion of helical swimmers (along their body axis) is non-reciprocal and can produce effective propulsion. As shown in **Fig. 7b**, the ABF contains a soft magnetic head (in yellow) and a helical tail (in gray). When the external magnetic field rotates either clockwise or counterclockwise, the ABF will be correspondingly actuated to swim forward or backward, respectively. In addition, the ABF swimmer can be driven to swim in new directions by changing the axis of rotation of the magnetic field. In recent years, bioinspired helical swimmers have become a classic design paradigm for low-Reynolds-number propulsion [185–189]. According to established research objectives, the presentation styles may vary in terms of size, methods of fabrication, structural design, and functionality.

Another representative of flagellar motility is inspired by sperms, swinging their slender flagella to form planar/helical chiral waves for propulsion. Khalil et al. propose a 322μm-long MagnetoSperm with a magnetically responsive head and a soft tail [190]. The cobalt-nickel layer provides a dipole moment that allows the MagnetoSperm to align the oscillating magnetic field to generate planar wave propulsion. As depicted in **Fig. 7c**, when controlled by a weak oscillating magnetic field (∼5 mT) at a frequency of 25 Hz, the MagnetoSperm advances 870 μm in 21 seconds. Bioinspired microswimmers with similar locomotion can be produced using various materials and processes, such as the DNA-linked chain of colloidal magnetic particles attached to a red blood cell for propulsion [58] and the biohybrid swimmer formed by a sperm coated with magnetic nanoparticles [191]. Moreover, controlling the magnetic field allows microswimmers to switch their swimming modes at will [192]. It is worth mentioning that these bioinspired microswimmers are significantly linked to recent advancements in biophysics [176,193–196]. As a result, any new discovery on the swimming tactics of microorganisms could help to improve the design and actuation methods of artificial microswimmers [197].

3.5.2 Millimeter-scale swimmers

Miniature soft magnetic robots are also developed with diverse swimming modes at the (sub)millimeter-scale. Ren et al. design a jellyfish-like soft magnetic robot, as shown in **Fig. 7d(i)**, which consists of a magnetic elastomer core and eight beating lappets, each with two compliant joints [53]. **Fig. 7d(ii)** illustrates the contraction and recovery phases of the robot's swimming under magnetic control, which involves changing the magnetic field direction and then strengthening the intensity. The lappets' proximal joints only bend in the recovery phase, while the distal joints bend in two phases. The forming non-reciprocal motion results in high thrust during the contraction phase and low resistance in the recovery phase, enabling the magnetic swimmer to swim efficiently in moderate Reynolds numbers. Diller et al. demonstrate a millimeter-scale soft magnetic swimmer with a classic magnetization profile [92] shown in **Fig. 7e(i)**. Each segment of the robot comprises distinct magnetization directions incorporated via a template-assisted magnetization method. Then, under the actuation of a rotating magnetic field, the robot can produce continuously propagating waves to propel itself forward on a water surface, as displayed in **Fig. 7e(ii)**.

As it can be seen above, the propulsion mechanism of miniature artificial swimmers is an enduring topic. Magnetic control, a well-controlled, non-invasive, and safe approach, can replicate various biomimetic swimming patterns and hold potential for biomedical applications, thus widely embraced by researchers.

## 3.6 Swarm or Assembly

Swarm behaviors, characterized by complex interactions and collaborations, are common in the biological world, from flying birds to ant groups [198,199]. Scientists have expressed great interest in how these behaviors arise, how they operate, and how they can be used [200–202]. For soft magnetic robots, their magnetic components can engage in diverse interactive behaviors and complete collaborative tasks under magnetic control. Here we briefly provide examples to discuss them (**Fig. 8**) [54,203–205].

Kuang et al. propose a unique concept that constructs morphing architectures by magnetic-assisted assembly [203]. As illustrated in **Fig. 8a(i)**, magnetic modules with specific magnetization directions are configured onto a fixed target under a weak magnetic field (3 mT). Then, they are welded at the jointing seam using an IR laser for about 30 s to obtain an integrated assembly structure (**Fig. 8a(ii)**). Like the motion in **Fig. 5c**, the assembled architecture can perform dynamic walking under the magnetic field control. **Fig. 8a(iii)** shows the simulation and experiment of walking locomotion. Further, when more modules assemble onto the architecture, it can deform in another style, as depicted in **Fig. 8a(iv)-(vi)**. The attractive thing is that the robot can continue to deform with new styles after in situ magnetization reprogramming. Through this novel design paradigm, magnetic modules can collaborate to establish diverse deformations or motions.

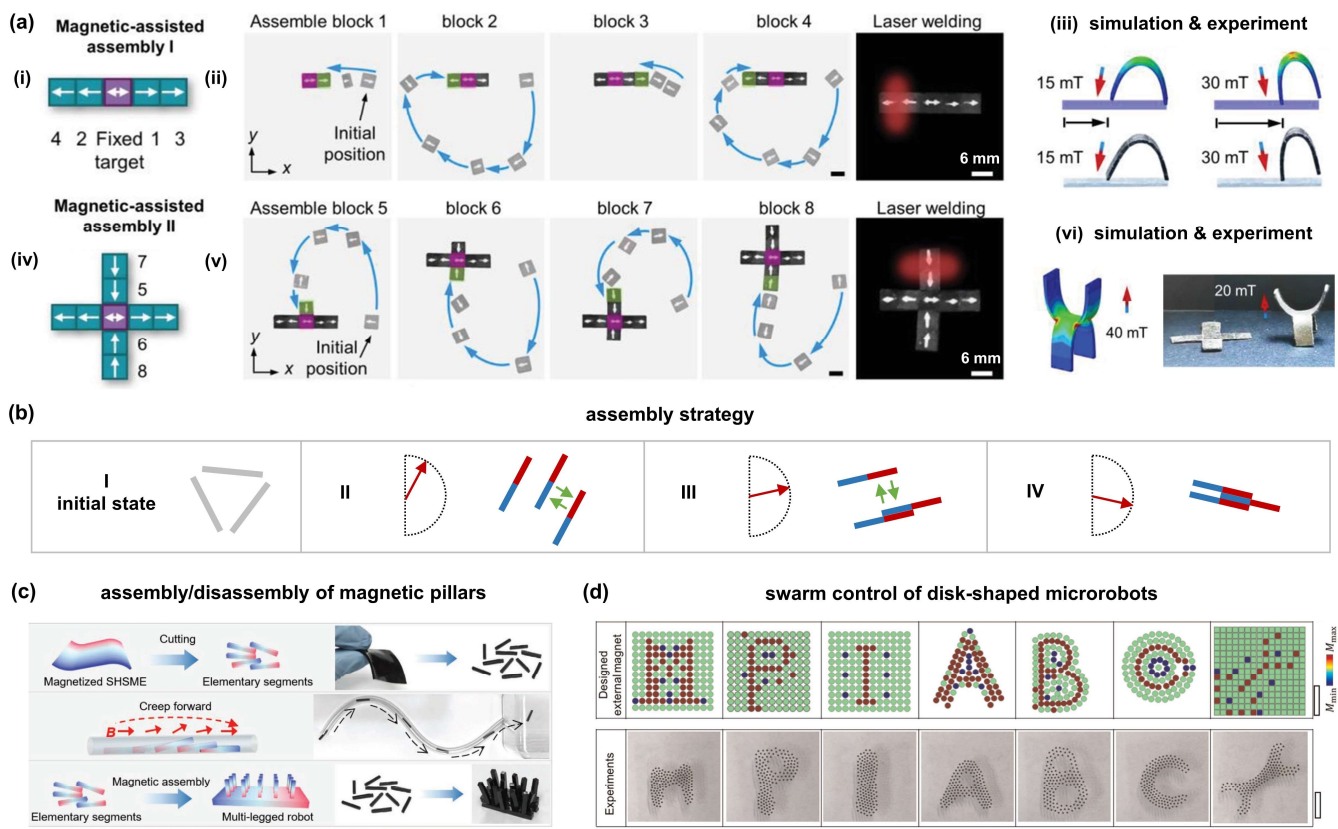

**Figure 8.** (**a**) Magnetic-assembled morphing architectures: (i)-(iii) the concept, assembly process, and locomotion of the mode I; (iv)-(vi) the concept, assembly process, and locomotion of the mode II (with permission from ref. [203], Copyright © 2021, Wiley-VCH). (**b**) Assembly strategy of needle-like microrobots. (**c**) Assembly and disassembly of self-healing supramolecular magnetic elastomers (with permission from ref. [54], Copyright © 2021, Wiley-VCH). (**d**) Magnetically controlled reconfiguration of microrobots: designed external magnets and experimental illustration (with permission from ref. [205], Copyright © 2020, SAGE Publishing).

Xu et al. present needle-like microrobots that can assemble through magnetic field control [204]. **Fig. 8b** illustrates the assembly mechanism. Each needle-like microrobot has its own magnetization direction. When the magnetic field direction changes, the "N" and "S" poles of adjacent microrobots attract each other to complete the assembly. With this method, microrobots can be assembled into one needle-like structure for further development. Cheng et al. propose self-healing supramolecular magnetic elastomers consisting of stick-shaped pillars that could be assembled and disassembled for performing complex tasks [54]. The elastomer is first cut into pillars that are small enough to pass through the "S" shape tube (**Fig. 8c**). After crawling over the tube (under the magnetic field actuation), they can build a multi-legged robot based on their magnetic assembly ability and fast self-healing property to perform subsequent tasks. Needle-like magnetic structures are also used to mimic artificial cilia that have abundant coordination behaviors. For example, metachronal motion is just a great representative of their collective cooperation [21,63,64,66,206]. Additionally, by programming the magnetic fields, reconfigured magnetic cilia carpet can switch its forms and complete multiple tasks, e.g., dexterous operations and controllable drug diffusion [207].

Dong et al demonstrate the swarm control of disk-shaped microrobots composed of NdFeB magnetic microparticles and casting resins [205]. With the well-designed arrangement of external magnets ($\phi$ 1 mm × 1 mm), 146 magnetic microrobots can form static formation patterns (including M/P/I/A/B/C/wrench-shaped formations) at the air–water interface (**Fig. 8d**). In these configurations, each microrobot has an upward magnetization direction, and the red dots (upper pole: N; lower pole: S) and blue dots (upper pole: S; lower pole: N) represent two types of arrangements of magnets.

Another representative example is the swarm control of paramagnetic nanoparticles, which completes the morphological reconstruction through the interaction among nanoparticles under the coupling effect of external magnetic fields and hydrodynamics [208]. This strategy, using tailored alternating magnetic fields to induce time-varying particle-particle interactions, has been widely studied in recent years [209–211]. However, like the disk-shaped microrobots [205], they rarely show the utilization of the soft property, i.e., beyond the scope of this review. We refer the reader to relevant reviews for more detailed information [32,48,212].

## 4. Application of Functionalized Soft Magnetic Robots

Having addressed the design, actuation, and biomimetic motion implementation of soft magnetic robots, the subsequent issue pertains to their functionalization and practical applications. This review primarily concentrates on the two most promising application domains for soft magnetic robots, i.e., their potential to revolutionize traditional robotic operations and biomedical applications [213,214]. Specifically, for small-scale manipulation tasks, we categorize them as solid manipulation (Section 4.1.1) and liquid operations (Section 4.1.2); with regard to biomedical applications, we introduce their utilization in therapeutic (Section 4.2.1) and diagnostic (Section 4.2.2), respectively.

*4.1 Small-Scale Manipulation Tasks*

4.1.1 Solid manipulation

Grasping and moving solid objects are typical robotic operations. Unlike traditional rigid robots, soft magnetic robots, inspired by biological behaviors, have been developed with novel approaches to accomplish various solid manipulation tasks. Here we introduce several representative soft magnetic robots/actuators that can perform grasping operations, cargo transportation, cargo sorting, and a series of solid manipulation tasks, as illustrated in **Fig. 9**.

As shown in **Fig. 9a(i)**, the Venus flytrap can catch its prey through a rapid closure movement, while displaying an open state in its natural situation. Inspired by this biological behavior, Zhang et al propose a soft magnetic gripper [215], as in **Fig. 9a(ii)**. The gripper consists of two bistable anti-symmetric shells and two neodymium magnets. The gripping motion can be triggered by applying the magnetic field to the magnets and finish in 0.112 s. Similarly, Wang et al demonstrate a flower-shaped, ultrafast soft magnetic robot composed of polydimethylsiloxane (PDMS) and NdFeB microparticles [140]. As in **Fig. 9b**, the soft magnetic robot with pre-magnetized directions performs grasping motion under an upward magnetic field control. The entire process can be completed within tens of milliseconds, which is less than the reaction time of a living fly (~100 ms). The experiment demonstrates the capture (at 35 ms) and release (at 56 ms) of a fly, and the result shows that it would not harm the fly's fragile body, as depicted in **Fig. 9b**. This is one of the typical advantages of soft robots' gripping operation, i.e., low damage and high adaptability. This gripping strategy has paved the way for enriching shape-morphing magnetic soft machines [216], grasping objects in different environments [217], and performing biomedical applications [218].

As a unique type of magnetic robot, liquid magneto-robots based on liquid metal or ferrofluids have been investigated to realize large deformations and show enhanced flexibility [219,220]. In **Fig. 9c**, Fan et al. present a ferrofluid droplet robot (FDR) to achieve a series of small-scale operations, including transporting and sorting cargo [20]. Cargo transport is facilitated by utilizing a ring-shaped permanent magnet to program the FDR. Specifically, (i) positioning the magnet parallel to the plane makes the upper FDR form a ring shape; (ii) inclining the magnet produces a C-shape FDR. By alternating between these custom shapes, the FDR is capable of collecting, transporting, and releasing cargo to reach the intended location. Moreover, the FDR can be actuated by a long cylinder-shaped magnet to sort cargo. When the force applied to the FDR exceeds the threshold, the FDR can break at the obstacles and recombine to perform subsequent operations. Such ability and flexibility do not exist in elastic-based soft robots. These liquid magneto-robots have recently demonstrated improved stability, and have been widely implemented for diverse applications, including small-scale robotic operations along with biomedical and electronic applications [221,222].

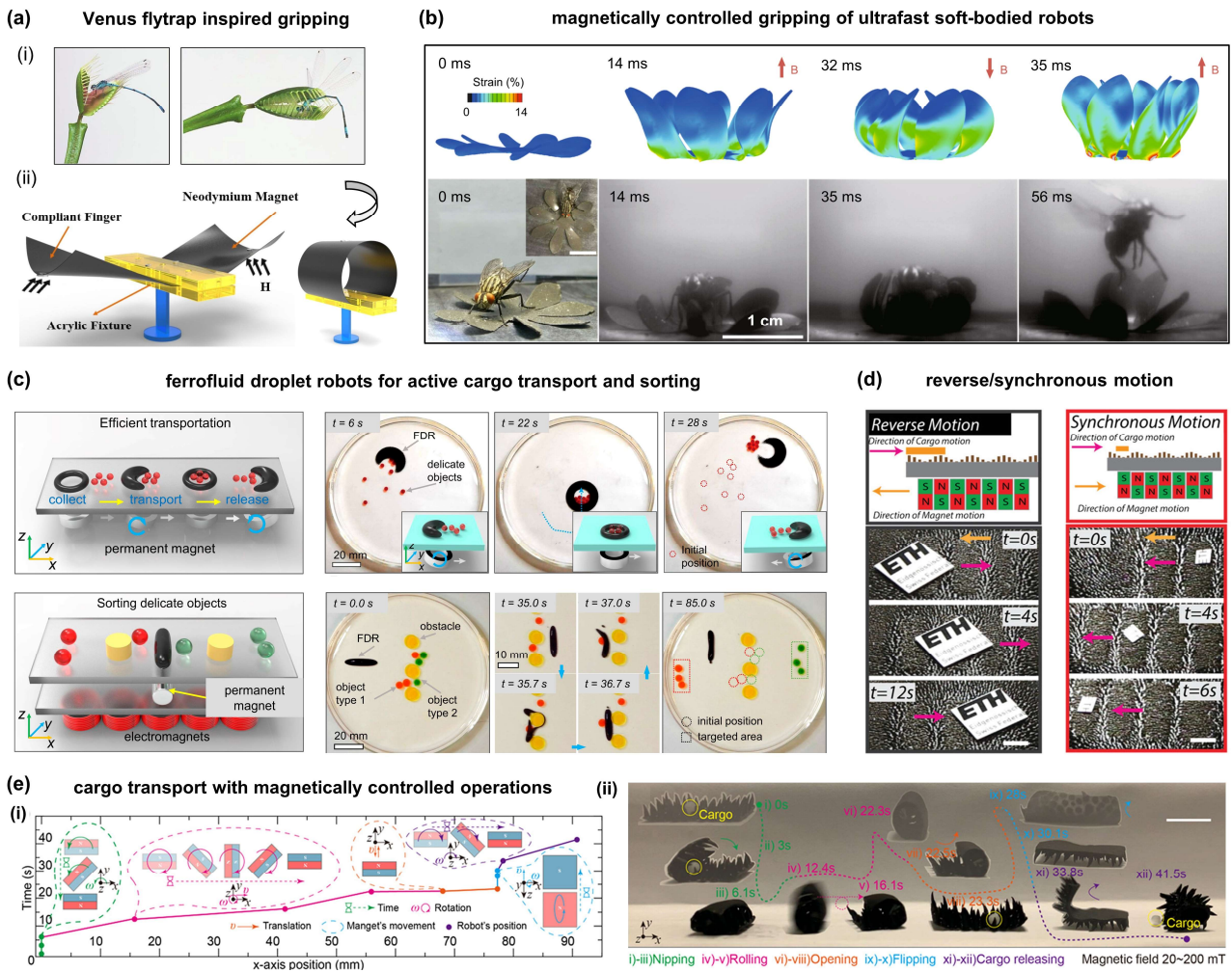

**Figure 9.** (**a**) Gripping motion of (i) the Venus flytrap and (ii) soft magnetic gripper (with permission from ref. [215], Copyright © 2019, Elsevier). (**b**) Gripping motion of the ultrafast soft magnetic robot (with permission from ref. [140], Copyright © 2020, Springer Nature). (**c**) Cargo transport and sorting of the FDRs (with permission from ref. [20], Copyright © 2020, National Academy of Sciences). (**d**) Reverse and synchronous cargo transport of the soft magnetic carpet (with permission from ref. [223], Copyright © 2021, Wiley-VCH). (**e**) Soft gelatin robot for targeted cargo transport and release: (i) magnetic control methods and (ii) experiments (with permission from ref. [78], Copyright © 2022, American Chemical Society).

Artificial cilia carpet is considered to be a versatile tool that can transport both solid objects and liquids, as suggested by several studies [21,63,65,66,223–225]. **Fig. 9d** shows the soft magnetic carpet's ability to transport cargo with various sizes [223]. During the experiments, the magnets' moving direction is the direction of the propagating metachronal wave. The cargo larger than a single magnet travels in the reverse direction as the magnet (i.e., reverse motion) due to the effect of metachronal coordination; however, the cargo smaller than a single magnet is trapped in the magnetic well, and then follows the magnets to move towards the same direction (i.e., synchronous motion). In addition, the multi-cilia structure can be utilized for gripping and transporting solid objects [78]. As the magnetic control methods are shown in **Fig. 9e(i)**, Yang et al. demonstrate targeted transport and release of cargo through a series of operations (including nipping, rolling, opening, flipping, and cargo release) using the soft magnetic robot, which is illustrated in **Fig. 9e(ii)**. This approach is similar to that of the previously mentioned soft magnetic grippers and ferrofluid droplet robots.

4.1.2 Liquid operation

Liquid operations are ubiquitous and critical to a great many practical applications [26,27,226,227], including microfluidics, heat management, chemical reactions, etc. The main strategies for liquid/droplet manipulation nowadays include designing surfaces with fixed asymmetric (such as meaning curved, inclined, and gradient) microstructures [228–231] or applying external field stimuli (e.g., light, electric, magnetic, and thermal stimuli) [232,233]. Recently, soft magnetic robots with arrayed microstructures, combining the advantages of physical structure design and external stimuli, have brought hope for highly controllable and flexible liquid operations. In this review, we illustrate the latest advances in this field by presenting several typical cases.

Metachronal waves are an extremely efficient method of propulsion in nature [234], and have been widely employed for manipulating liquids. As illustrated in **Fig. 10a**, Dong et al. present bioinspired cilia arrays that can perform various liquid operations when actuated by magnetic fields [21]. Propelled by metachronal waves, viscous fluids can be transported on curved surfaces towards the opposite direction of propagating waves (**Fig. 10a(i)**). To build a local mixing region, the researchers arrange symplectic metachronal waves (power stroke and propagating wave direction are the same) on the top array and antiplectic metachronal waves (power stroke and propagating wave direction are opposite) on the bottom array. The mixing process at different times is illustrated in **Fig. 10a(ii)**. Benefiting from the unique beating modes of natural cilia, such artificial cilia are suitable for operating microscale fluids, which are dominated by viscous forces, including but not limited to liquid transport and mixing [65,223,235,236].

Jiang et al present a multifunctional liquid manipulator that is composed of microplate arrays with vertically aligned magnetic particles [237]. By modifying the surface with a superhydrophobic spray, the manipulator becomes increasingly hydrophobic and less adhesive, as shown in **Fig. 10b(i)**. Consequently, magnetically responsive microplates can transport a water droplet in a metachronal manner (**Fig. 10b(ii)**). Relying on the adhesion force, the magnetic microplates array can also vertically capture the water droplet. When two jointed magnets are set above the microplates array, the droplet can be controllably released by the structural bending, as illustrated in **Fig. 10b(iii)**. This study combinedly exploits the physical structures of surfaces and external energy fields, which further expands the potential of soft magnetic robots in liquid operations. Recently, inspired by pine needles and cilia, Miao et al propose a magnetic pillar actuator that can controllably transport liquids [63]. The asymmetric-structured design allows liquids to be transported directionally on a static surface (**Fig. 10c(i)**). As the pillar arrays present a gradient bending deformation (controlled by magnetic fields), the liquid transportation process can be increased in speed from 38 s to 15 s (the directions of static liquid transport and gradient reduction are the same) or changed to another direction (gradient reduction direction), as shown in **Fig. 10c(ii)**. This strategy, inducing the multiple biomimetic designs into magnetic robots, enables them to support the integration of rich functions, e.g., liquid transport and robotic locomotion.

Some recent studies have focused on revolutionizing the paradigm of liquid/droplet operations by using magnetic robots across scales (from micro/nano-scale to millimeter-scale) [21,63,237–240]. Such an exciting interdisciplinary topic may benefit many research fields, involving soft robotics, smart materials, microfluidics, surface science, and other engineering/medicine issues.

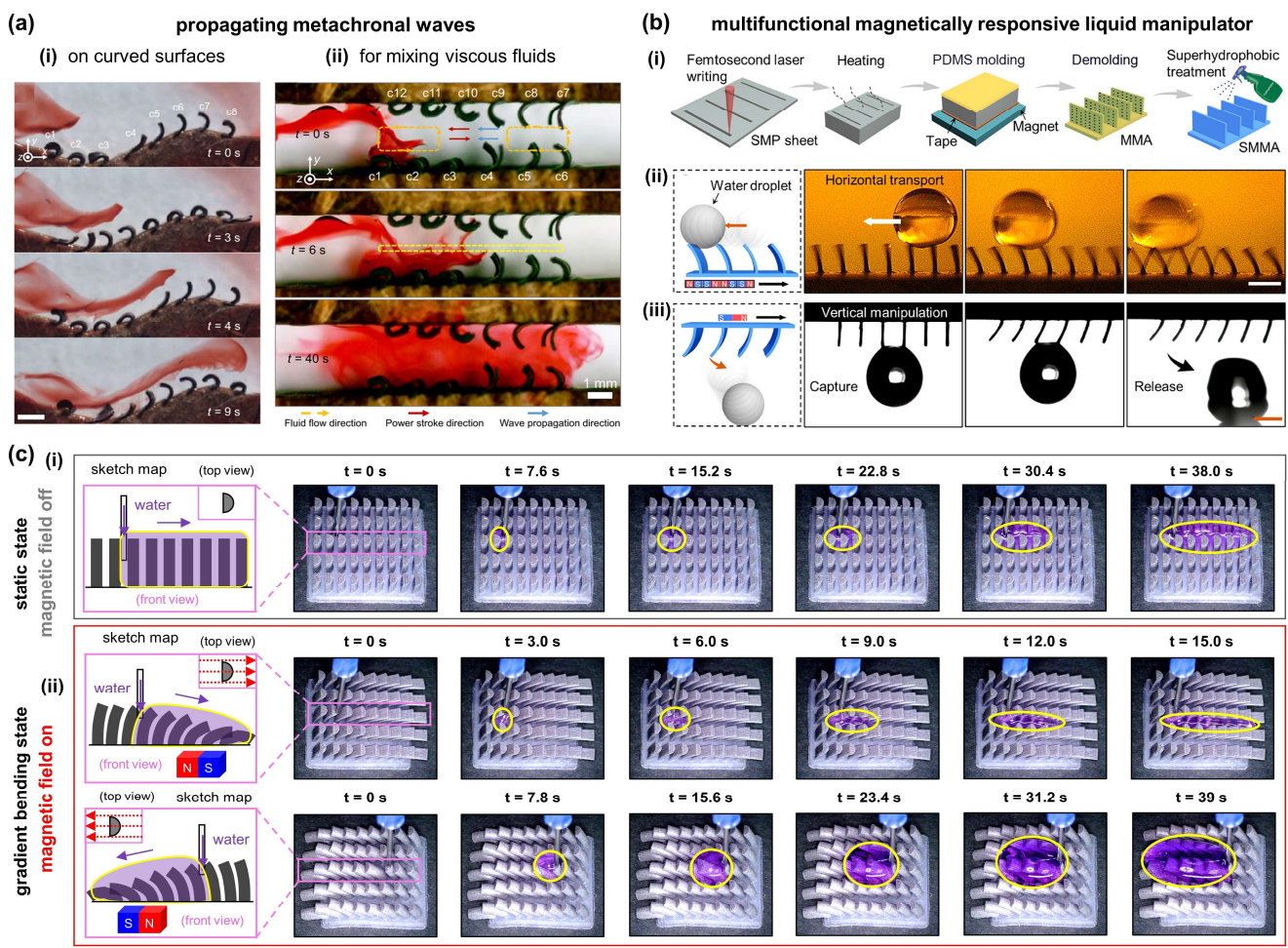

**Figure 10.** (**a**) The metachronal motion of bioinspired magnetic cilia arrays: (i) on curved surfaces; (ii) for liquid mixing (with permission from ref. [21], Copyright © 2020, American Association for the Advancement of Science). (**b**) Multifunctional magnetic liquid manipulator: (i) fabrication; used for (ii) water droplet transport and (iii) water droplet capture and release (with permission from ref. [237], Copyright © 2020, American Chemical Society). (**c**) (i) Liquid transport of the static magnetic pillar actuator. (ii) Magnetically controlled liquid transport with an enhanced speed or a changed direction (with permission from ref. [63], Copyright © 2022, American Chemical Society).

## *4.2 Biomedical Applications*

4.2.1 Therapeutic use

The long-term vision of miniature magnetic robots is to carry out various medical operations (from cellular to organ level) in the human body to replace current invasive surgeries [22]. While magnetic shape-morphing micro/nanorobots have been proposed [241], soft magnetic robots that can highly rely on abundant deformations/motions to overcome obstacles within the human body and provide minimally invasive surgery or targeted drug delivery, are mainly concentrated at the (sub)millimeter to centimeter scale. Consequently, this review is focused on various types of biomedical applications of the (sub)millimeter/centimeter-scale soft magnetic robots. The latest developments in medical micro/nano-scale magnetic robots have been well examined in recent reviews [49,242–245]. Hence, we suggest the reader to find more information in them.

The soft magnetic robot with multimodal motions illustrated in **Fig. 11a** has the potential to perform medical tasks in the gastrointestinal (GI) tract [18]. To achieve this goal, the primary condition is to enable itself to overcome various obstacles in the complex gastrointestinal environment and reach the target location. Researchers have shown the multimodal locomotion of the soft robot in a hybrid liquid–solid environment to display its strong motion ability (**Fig. 11a(i)-(v)**). The soft robot can freely adapt to water and land environments, immerse itself in/out of the water, cross obstacles, and pass through the narrow pipe by effortlessly switching different locomotion modes. Lastly, the robot can finish the navigation in a synthetic stomach phantom via a combination of jumping, rolling, climbing, and landing locomotion. Also, the robot's potential for integration with clinical surgeries is explained through the ultrasound-guided rolling motion displayed in an *ex vivo* chicken tissue (**Fig. 11a(vi)**).

Currently, identifying appropriate medical scenarios for soft magnetic robots is a big challenge. Soft magnetic robots' *in vivo* medical operations are restricted by real-time imaging, precision control, and security issues. Consequently, the clinical application of these soft magnetic robots is impeded, and they are still in the lab testing stage. Xu et al develop a soft magnetic robot for the non-invasive crushing of gallstones [246], as depicted in **Fig. 11b(i)-(ii)**. The robot is spherical with an average diameter of 5 mm, and has a spiny skin with local microscale features, which can prevent adhesion to the digestive tract and increase pressure on collisions with gallstones. The surgical process is shown in **Fig. 11b(iii)-(vi)**. Controlled by magnetic fields, the soft robot can move through organs and tissues into the gallbladder. Following this, the application of a pulsed magnetic field leads to the collision of the robot and gallstones. Eventually, the gallstone residue is excreted with the contraction of the gallbladder, and the robot made of ferromagnetic powder and flexible silicone rubber materials shows good biocompatibility.

In addition to non-invasive surgery in the human body, targeted drug delivery is another classic therapeutic biomedical operation. Zhang et al have proposed a paradigm for the oral delivery of macromolecular drugs by combining magnetic robots with microneedle technology [55]. As the concept illustrated in **Fig. 11c**, the robot contains a magnetic substrate, a separable connection, and microneedles loaded with drugs. The microneedle robot is encapsulated into a commercial enteric capsule to ensure its stability in gastric juices and smooth release in the small intestine. Driven by the magnetic force, the robot's microneedle part can penetrate the intestinal barrier and insert into the tissue to deliver drugs. Then, the separable connection part, composed of a low concentration of gelatin methacryloyl mixed with bovine serum albumin, can be quickly degraded by the enzymes, allowing the magnetic part to be expelled from the body and leaving the drug-loaded tips in the tissue. The insulin-loaded magnetic microneedle robot has been administered orally to pigs, proving its efficacy in effectively controlling their blood glucose levels. Similar needle insertion operations via magnetic forces [247] or magnetically actuated mechanisms [248] have shown the potential for targeted drug delivery and surgery.

To conclude, several actual factors restrict the biomedical applications of miniature soft robots, including (i) the complex and narrow three-dimensional environment within the human body; (ii) dampness, viscosity, and other characteristics of organs/tissues; (iii) biological barriers, such as blood–brain barrier, mucosa, and mucus; (iv) imaging limitations; (v) the robot's robustness, intelligence and safety. Although some magnetic robots have been proposed to tackle these challenges [249,250], the stability and safety of their clinical applications still need to be evaluated and improved.

4.2.2 Diagnostic use

Miniature soft magnetic robots have the potential to aid in diagnostic tasks, including disease examination and detecting physiological signals. Despite their superior motion performance, soft magnetic robots need to be integrated with imaging, sampling, or other specific sensing capabilities to successfully aid in biomedical detections and diagnoses. Here, through examples of the magnetic robotic biopsy endoscope and magnetic sensing robot, we illustrate the significance of combining the robotic deformations/motions with advanced sampling and sensing capabilities for disease examination and early warning.

Son et al. propose a magnetically controlled soft capsule robot for biopsy under the surface of the GI tract [60]. The soft robot is encapsulated in the ice capsule to prevent it from causing damage to the esophagus. Then, relying on the rolling motion and a carrying camera, the magnetic robot can serve as a visual endoscopy and reach the suspicious lesion location. The robot includes a magnet, a fine needle, a camera, a tether, and other components. Actuated by the magnetic force, the robot can collapse and expose the biopsy needle, whereas the magnetic torque is used to stabilize the robot's orientation angle. After the biopsy, the robot can be retracted by a tether attached to the robot's top part. During the *in vitro* test of the porcine stomach, the robot shows a biopsy yield of 85%.

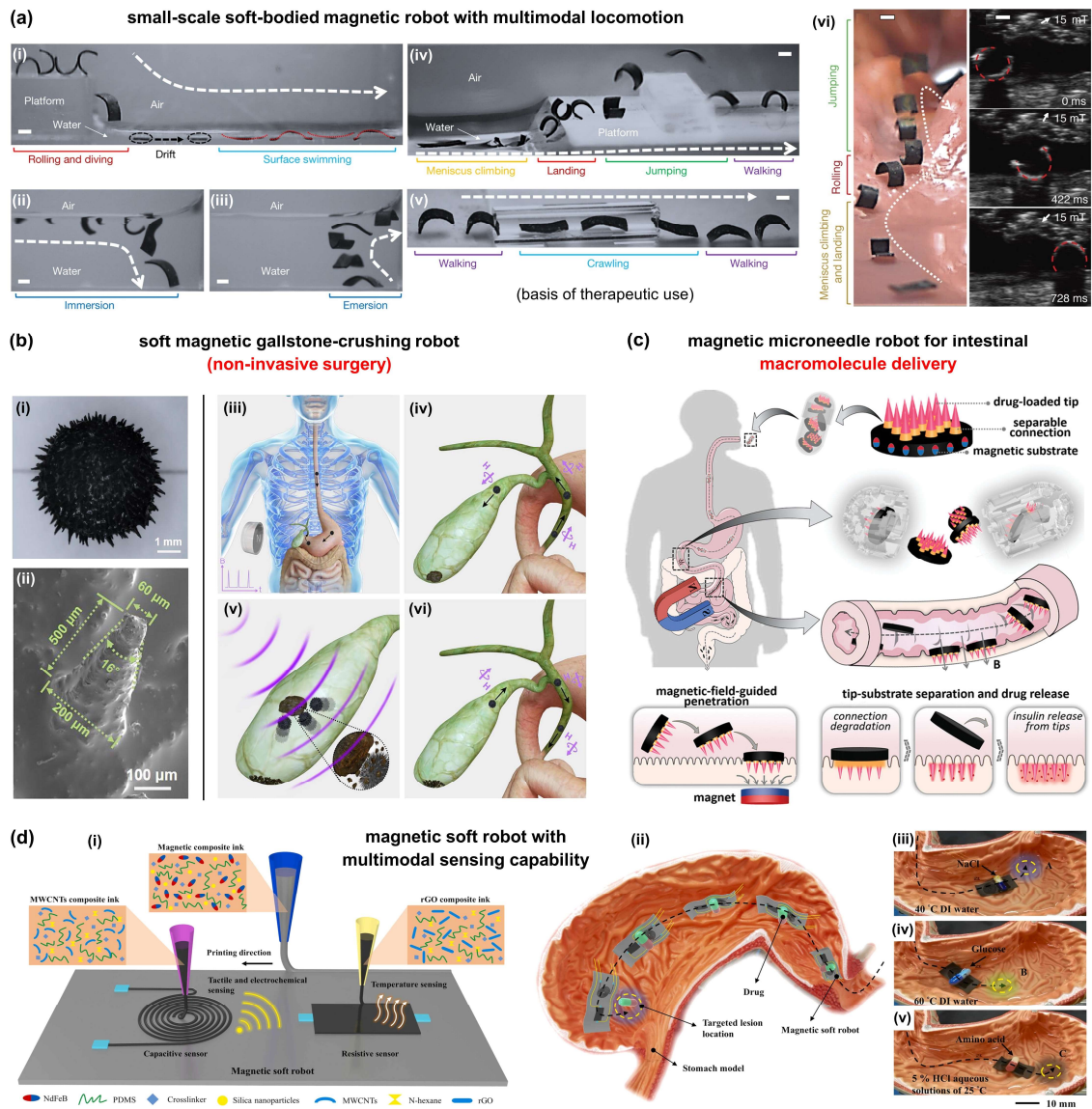

**Figure 11.** (**a**) Multimodal locomotion of the soft robot in complex environments: (i)-(v) a hybrid liquid–solid environment and (vi) biomedical scenarios (with permission from ref. [18], Copyright © 2018, Springer Nature). (**b**) Soft magnetic gallstone-crushing robot and its lithotripsy mechanism (with permission from ref. [246], Copyright © 2022, Elsevier). (**c**) Illustration of the magnetic microneedle robot for intestinal macromolecule delivery, including the composition, actuation, and drug delivery mechanism (with permission from ref. [55], Copyright © 2021, Wiley-VCH). (**d**) (i) Design of the magnetic soft robot with multimodal sensing abilities. (ii) Illustration of monitoring targeted drug delivery experiments. (iii)-(v) Three types of drug delivery experiments of the soft magnetic robot (with permission from ref. [56], Copyright © 2023, Elsevier).

Wang et al. develop a soft magnetic robot through a multi-material direct ink writing process [56], which holds the temperature, tactile, and electrochemical sensing capabilities, as shown in **Fig. 11d(i)**. The soft magnetic robot contains two sensing parts: one is a double-involute capacitive sensor, composed of multi-walled carbon nanotubes (MWCNTs) and PDMS, for tactile and electrochemical sensing; the other is a rectangular resistive sensor, composed of reduced graphene oxide (rGO) and PDMS, for temperature sensing. The increased temperature makes the activated charge hopping of localization states, causing the resistance decrease of the resistive sensor. With this sensing mechanism, the resistive sensor can monitor the changes as low as 0.5 °C of human temperature for the biomedical operations of the magnetic robot. Moreover, based on the ideal parallel plate capacitance theory $C = \varepsilon\varepsilon_0 S/d$, the tactile stimuli of different materials (e.g., iron, stone, wood, and glass) can change the capacitance response by changing $\varepsilon$ (relative dielectric permittivity) or $d$ (effective distance). Also, because the fringing capacitances of the arch-shape sensing elements are sensitive to the $\varepsilon$ of the chemical solutions, researchers use it to quantitatively evaluate the soft magnetic robot's three targeted drug delivery tests (**Fig. 11d(ii)-(v)**), i.e., (i) deliver the NaCl capsule to 40º DI water (location "A"); (ii) deliver the glucose capsule to 60º DI water (location "B"); (iii) deliver the amino acid capsule to 5% HCl aqueous (location "C"). In the experiments, the capacitance value stabilizes again after hundreds of seconds, meaning that the drug is fully delivered.

In summary, the promising directions for diagnostic magnetic soft robots are concentrated in the above two categories. Firstly, the capsule robots, which are currently relatively mature [251,252], can ensure accurate disease examination with both physical monitoring (imaging) and physiological sampling (biopsy). Secondly, with the support of advanced sensing technologies [253], achieving a soft medical robot that integrates diagnosis and treatment is an important goal for future medical robots.

## 5. Conclusions and Perspectives

Soft magnetic robots play an important role in miniature soft robotics considering their great flexibility and rich functionality [4,24,254,255]. This review summarizes the different magnetic materials and their suitable fabrication methods. The representative actuation systems are sequentially delivered, from permanent magnets to electromagnetic coils. Their emerging applications in small-scale manipulation and biomedical fields are discussed. These soft robots have made significant progress in material processing, design concepts, control methods, and wide application scenarios, while further efforts are still needed to enhance intelligence, functionality, control robustness, and security. In the meantime, the cross-field applications extended by soft magnetic robots has been a hot research trend in recent years, which deserves our attention. Here we discuss underexplored areas of soft magnetic robots and suggest directions for their future developments. The perspectives are concluded as below.

- *Advanced smart materials*: The rising of novel materials not only opens up new possibilities for soft magnetic robots in typical robotic tasks (such as grasping, locomotion, and dexterous operations), but increases previously unimaginable capabilities (e.g., biodegradability and self-healing). The development of intelligent soft matter can continuously bring new application paradigms and unprecedented opportunities [256–259]. In recent years, a continuous effort is to construct biocompatible or biodegradable magnetic microrobots [260,261]. These studies often use hydrogel-based materials, combined with advanced 3D printing technology (e.g., two-photon polymerization [262]), to achieve massive and standardized production of magnetic robots. Robotic safety is often seen as the primary issue in biomedical applications because it ensures the fundamental usability of robots. Furthermore, soft materials with self-healing properties endow soft magnetic robots with excellent environmental adaptability and mechanical durability [54]. Multi-physical actuation (by coupling magnetic actuation with other actuating strategies, e.g., light, electricity, chemistry, humidity, and temperature) is another strategy to construct increasingly functional and intelligent soft robots [263]. The requirement to benefit from this strategy is that the effects of multi-physical actuation are independent of each other and even can complement each other's shortcomings. Also, the subsequent trade-off between system complexity and versatility needs to be carefully considered. Addressing these concerns depends largely on the development and synthesis of novel smart materials.

- ***Design and fabrication technology***: Numerical solutions based on force and moment balances can be employed to find the optimal magnetization profile for programming the desired shape in simple geometries like beams or rods. But when it comes to complicated geometries, it will be a tough task to find the required magnetization patterns for realizing desired shape changes. Except for trial and error, machine learning has been shown as a powerful tool to solve inverse-design problems for functional composites [264]. It is foreseeable that in future research, how programmable magnetic robots can be designed in an innovative and efficient manner is a key issue. Further, how to develop low-cost, high-precision, powerful manufacturing technology that can fabricate complex structures is also a main point to be considered.

- ***Robot control and integration***: Most of the current soft magnetic robots are still open-loop control systems. Utilizing emerging hardware and algorithms [265,266], the integration of closed-loop control, self-sensing, and decision-making enables magnetic soft robot systems to perform complex tasks. In the future, soft magnetic robots, which are integrated with self-powered components and capable of real-time feedback and modulation, are highly expected.

- ***Swarm and computational intelligence***: Operating multiple robots that work closely together is an appealing research direction [267,268], which brings new abilities and high efficiency to magnetic robots. Such group coordination appears to be across scales and may be based on different formation mechanisms. Additionally, the introduction of computational intelligence for soft magnetic robots is attractive. It enables each module or motion mode of the robot to demonstrate autonomous adaptation to changes in unknown and unstructured environments and make appropriate decisions [269–271].

- ***Multi-functionality and versatility***: Inspired by nature, soft magnetic robots have shown a variety of biomimetic motions and have been endowed with attractive functions, facilitating their smooth execution of small-scale manipulation tasks and biomedical applications. Recently, it has been proved that magnetic robots with on-board sensing systems can enable controllable assembly processes, paving the way for electronic devices with self-assembly or autonomous rearrangement abilities [75,272]. Miniature robots integrated with sensing modules enable intelligent feedback-controlled actuation [273], and the demonstrated research motivation of these studies will indeed accelerate the magnetic robots' real-world applications. In addition, the cross-field applications extended by soft magnetic robots deserve more attention. The concept of magnetic robots can be defined widely. It may refer to all the magnetically responsive active systems. Thus, this active and automated concept can be integrated into many microsystems (involving multiple research fields) to enrich the functionality of magnetic robots, e.g., adaptive soft magnetic stents [274] and magnetic-controlled automated virus detection platforms [275].

- ***Security***: To reach clinical applications, soft magnetic robots are facing challenges in materials design, precise control, and safety problems [24]. The environment inside the human body is extremely complex, often causing the failures of wireless surgeries and disease examinations of soft magnetic robots. Meanwhile, a series of safety issues remain unsolved, for instance, how to ensure human security in the robotic surgery and examination process, and if the robots can be smoothly recycled or safely degraded. Solving these problems need highly interdisciplinary approaches and close cooperation among materials scientists, engineers, biophysicists, and clinicians.